\documentclass[fleqn,10pt]{wlscirep}
\usepackage[utf8]{inputenc}
\usepackage[T1]{fontenc}
\usepackage{amsfonts,amssymb}
\usepackage{multirow}
\usepackage{graphicx}
\usepackage[figuresleft]{rotating}
\usepackage{ulem}
\usepackage{xcolor}
\title{Enhancing Source Code Classification Effectiveness via Prompt Learning Incorporating Knowledge Features}

\author[1,3]{Yong Ma}
\author[1]{Senlin Luo}
\author[2,*]{Yu-Ming Shang}
\author[1]{Yifei Zhang}
\author[1]{Zhengjun Li}
\affil[1]{Beijing Institute of Technology, Beijing, 100085, China}
\affil[2]{Beijing University of Posts and Telecommunications, Beijing, 100876, China}
\affil[3]{Qi-Anxin Technology Group, Beijing, 100044, China}

\affil[*]{shangym@bupt.edu.cn}

\begin{abstract}
{	
Researchers have investigated the potential of leveraging pre-trained language models, such as CodeBERT, to enhance source code-related tasks. Previous methodologies have relied on CodeBERT's '[CLS]' token as the embedding representation of input sequences for task performance, necessitating additional neural network layers to enhance feature representation, which in turn increases computational expenses. These approaches have also failed to fully leverage the comprehensive knowledge inherent within the source code and its associated text, potentially limiting classification efficacy.
We propose CodeClassPrompt, a text classification technique that harnesses prompt learning to extract rich knowledge associated with input sequences from pre-trained models, thereby eliminating the need for additional layers and lowering computational costs. By applying an attention mechanism, we synthesize multi-layered knowledge into task-specific features, enhancing classification accuracy. Our comprehensive experimentation across four distinct source code-related tasks reveals that CodeClassPrompt achieves competitive performance while significantly reducing computational overhead.
}
\end{abstract}
\begin{document}

\flushbottom
\maketitle
%
%
\thispagestyle{empty}


\section*{Introduction}

\label{intro}



The intersection of machine learning, programming languages, and software engineering has garnered significant interest from the software engineering community.
{The classification of source code-related tasks holds significant importance in the field of software engineering, as it serves the purpose of identifying the programming language used in the source code and enhancing the overall software quality. Many techniques, including Bayesian methods \cite{khasnabishDetectingProgrammingLanguage2014}, Random Forest \cite{breimanRandomForests2001}, and XGBoost \cite{alrashedySCCPredictingProgramming2020a}, have been employed to accomplish source code-related classification. Additionally, deep learning techniques, such as TextCNN \cite{gildaSourceCodeClassification2017b}, have gained substantial prominence and are extensively utilized in this domain.}
Recently, large-scale pre-trained language models, such as BERT \cite{devlinBERTPretrainingDeep2019}, RoBERTa \cite{liuRoBERTaRobustlyOptimized2019}, GPT \cite{radfordImprovingLanguageUnderstanding2018}, and T5 \cite{raffelExploringLimitsTransfer2020a}, have emerged as promising tools for various downstream Natural Language Processing (NLP) tasks \cite{qiuPretrainedModelsNatural2020a}. The impressive performance of pre-trained language models (PLM) in NLP tasks has inspired intensive research in the field of software engineering. Source code-related tasks have achieved remarkable success with source code-dedicated pre-trained language models, such as CodeBERT \cite{fengCodeBERTPreTrainedModel2020b}, CodeT5 \cite{wangCodeT5IdentifierawareUnified2021}, and GraphCodeBERT\cite{guoGraphCodeBERTPretrainingCode2021}, as demonstrated by recent studies \cite{kwonCodeBERTBasedSoftware2023, kanadeLearningEvaluatingContextual2020}.

{
The CodeBERT model, a member of the BERT family, is extensively utilized in the field of software engineering\cite{fengCodeBERTPreTrainedModel2020b}. In BERT family models, the '\texttt{[CLS]}' token functions as the classification token and serves as the aggregate representation of the entire input sequence\cite{devlinBERTPretrainingDeep2019}. Previous research consistently adopts this design choice, employing the representation of the '[CLS]' token for various classification tasks\cite{choiEvaluationBERTALBERT2021,devlinBERTPretrainingDeep2019}. It leverages the highest layer of vector output of the '\texttt{[CLS]}' token from a BERT family model as the representation for an input sequence. However, relying solely on the '\texttt{[CLS]}' token in the highest layer of model output as the representative of the entire input text sequence imposes limitations on the effectiveness of the BERT family. Previous research \cite{goldbergAssessingBERTSyntactic2019, jawaharWhatDoesBERT2019} has demonstrated that BERT-based models can capture a rich hierarchy of linguistic information in NLP tasks. Specifically, surface features are captured in lower layers, syntactic features in middle layers, and semantic features in higher layers \cite{jawaharWhatDoesBERT2019}. Other studies propose utilizing the '\texttt{[CLS]}' tokens from multiple layers of vector output to more comprehensively represent the input information \cite{liuELCodeBertBetterExploiting2022}. Nonetheless, relying solely on the '\texttt{[CLS]}' token from the output of a BERT model as the exclusive representation of an input text is considered insufficient \cite{choiEvaluationBERTALBERT2021}. Some investigations have incorporated supplementary neural network layers, such as Long Short-Term Memory (LSTM) networks \cite{hochreiterLongShortTermMemory1997a}, to enhance the capabilities of feature representation. However, it is important to note that this approach comes with a notable increase in computational overhead, including the movement and computation of additional parameters.
}

{
The reason behind the use of '\texttt{[CLS]}' tokens in combination with LSTM is the inadequate capacity for feature representation on source code related tasks.
 Fortunately, the rapid development of large-scale language models \cite{brownLanguageModelsAre2020} is bringing promising solutions by introducing prompt learning as the fourth paradigm for both programming and natural language processing tasks \cite{liuPretrainPromptPredict2021}. 
During prompt learning, the '[MASK]' token can aggregate abundant knowledge associated with input sequences, including source code. By leveraging knowledge that goes beyond surface, syntactic, and semantic features, the use of LSTM layers can be rendered unnecessary, leading to reduced computation costs and improved feature representation.
}

{
The approach to leveraging prompt learning is to use a natural language prompt template to wrap the input text sequence, followed by performing masked language modeling with a pre-trained language model. For instance, in a text classification task, the text sequence $\mathbf{x}$ can be wrapped into a prompt template, such as "It was [MASK]. $\mathbf{x}$," and then input into a language model. The logits derived from the language model's output contain a wealth of retrieved knowledge at the location of '[MASK]', which can be mapped to a specific category for classification tasks.
}

In the domain of software engineering, the pursuit of a novel classification approach that offers low computational costs and possesses a robust capability for feature representation has emerged as a pivotal necessity.
To enhance the performance of tasks related to source code processing while minimizing computational costs, a promising approach is to leverage the knowledge extracted from a language model through prompt learning to represent the input source code or related text sequence. BERT-based models, such as CodeBERT, produce multiple layers of output, where each layer's output at the '[MASK]' position encapsulates specific knowledge relevant to the input information. By consolidating the important knowledge conveyed by the language model for each task, we can obtain a more comprehensive and discriminative feature set, {thereby  improving the accuracy of the task without the need for additional feature-enhancing mechanisms.}



{
Motivated by the aforementioned concepts, we propose a novel approach, CodeClassPrompt, to enhance the effectiveness of source code classification through prompt learning that incorporates knowledge features. The CodeClassPrompt leverages the comprehensive knowledge output from CodeBERT, a bimodal language model with the ability to process both natural language and programming languages.}
Our method revolves around the utilization of a carefully crafted prompt template, which encapsulates the input source code or related text. Subsequently, we employ a language model to extract knowledge that encapsulates the intrinsic characteristics of the input text. By leveraging the attention mechanism\cite{vaswaniAttentionAllYou2017a}, we aggregate the significance of knowledge across various layers for a given task and map it to a specific class. This approach enables us to fully exploit the extensive knowledge embedded in each layer of the BERT-based model through prompt learning, obviating the need for additional neural network layers for feature extraction and reducing computational costs. The aggregation of pivotal knowledge through the attention mechanism heightens the effectiveness of feature representation, thereby enhancing task accuracy.

{
To validate the efficacy of our proposed approach, we conducted experiments on four representative downstream tasks in the realm of source code: code language classification, code smell classification, technical debt classification, and code comment classification. The experimental results validate that our method attains comparable performance to prior studies for all four tasks, while concurrently reducing computational costs. Furthermore, we performed ablation experiments to assess the reliability of our components across various task configurations.
}

To the best of our knowledge, the main contributions of our study can be summarized as follows:
\begin{enumerate}
	
	\item Our paper is the first to combine the prompt-learning paradigm with source code-related tasks, advancing the technical progress of the software engineering field.
	
	\item { We present CodeClassPrompt, a pioneering approach that leverages the knowledge features acquired through prompt learning and aggregates indispensable knowledge using an attention mechanism. As a consequence, our study yields new state-of-the-art results on the Code Smell dataset and comparable results on three other datasets compared to previous investigations, while substantially mitigating computational costs.}
	
	\item We evaluate our approach on four classical source code-related tasks and demonstrate its effectiveness on both programming language and natural language tasks through a comprehensive set of experiments.

	\item  We have made our trained models and related code publicly available in our GitHub repository\footnote{https://github.com/BIT-ENGD/codepromptclass} to facilitate researchers in reproducing the results of our study or conducting further research.
	
\end{enumerate}

This paper is organized as follows. Section "Related Work" provides an overview of the relevant literature concerning CodeClassPrompt. Section "CodeClassPrompt" presents a comprehensive description of the design of our proposed approach. In Section "Experimental Study", we delineate the experimental setup of CodeClassPrompt and provide detailed information about the baselines employed in the experiments. Section "Results and Analysis" presents the experimental results and offers a thorough analysis of the findings. Moreover, we address potential threats to the validity of our approach in Section "Threats to Validity". Finally, we conclude the paper in Section "Conclusion".

\section*{Related Work}
\label{sec:1}


\subsection*{Code Representation Learning}

Significant advances have been made in the study of the intersection of machine or deep learning, programming languages, and software engineering, based on the assumption that programming code resembles natural language \cite{allamanisSurveyMachineLearning2018}. However, raw source code cannot be directly fed into machine or deep learning models, and code representation is a fundamental step in making source code compatible with these models. This involves preparing a numerical representation of the source code that can be used to solve specific software engineering problems.


In this subsection, we briefly introduce representative studies related to source code representation learning. As source code has rich structure, it cannot be treated as only a series of text tokens \cite{allamanisSurveyMachineLearning2018,nguyenDeepNeuralNetwork2018}. 
Harer et al. \cite{harerAutomatedSoftwareVulnerability2018a} tokenize a code snippet and categorize all tokens into different bins, such as comments and string literals. All tokens with the same categorical information are mapped to a unified identifier, which is then transformed into a vector using the word2vec algorithm \cite{leDistributedRepresentationsSentences2014a}. 
DeFreez et al. \cite{defreezPathbasedFunctionEmbedding2018} proposed the Func2Vec method, which embeds the control-flow graph of a function as a vector to represent the function and facilitates estimating the similarity of functions. 
A context-incorporating method \cite{nguyenDeepNeuralNetwork2018} was proposed to use syntactic and type annotations information for source code embedding, which can distinguish the lexical tokens in different syntactic and type contexts. 
Code2Vec \cite{alonCode2vecLearningDistributed2019} offers a new approach, which decomposes code into a collection of paths in its abstract syntax tree (AST), learning the atomic representation of each path and how to aggregate a set of them. 
For the problem of large ASTlength, Zhang et al. \cite{zhangNovelNeuralSource2019} found that splitting a large AST into a number of small statement trees and then encoding them as vectors can capture both lexical and syntactic knowledge of the statements. 
Motivated by the need for code summarization, Hu et al. \cite{ijcai2018p314} proposed a simple representation of code that only leverages the vectors of a sequence of API names to express a code snippet. 
{With the advancement of PLMs, a number of methods based on PLMs have been proposed. Yang et al. \cite{yangFinegrainedPseudocodeGeneration2021} offer a fresh way by utilizing a PLM and convolutional neural networks (CNN) to extract the feature representation of a code snippet. 
Since BERT is not dedicated to source code-related tasks, Feng et al. \cite{fengCodeBERTPreTrainedModel2020b} utilize both code snippets and natural language to construct a bimodal pre-trained model, CodeBERT, that can effectively represent source code-related material. 
Recently, Jain et al. \cite{jainContrastiveCodeRepresentation2021} introduced contrastive learning into code representation learning, as BERT-based models are much more sensitive to source code edits and cannot represent similar code snippets with slightly different literal expressions.}


\subsection*{Source Code Related Classification}

Source code-related classification, which includes code language classification, code smell classification, code comment classification, and technical debt classification, are four crucial tasks in software engineering that have been thoroughly investigated by researchers.

\subsubsection*{Code Language Classification}

In most source code-related tasks, code language classification is the initial step for further processing. Previously, the programming language of a piece of source code was assigned manually or determined based on its file extension \cite{gildaSourceCodeClassification2017b}. 
SC++ \cite{alrashedySCCPredictingProgramming2020a} employs the Random Forest Classifier (RFC) and XGBoost (a gradient boosting algorithm) to build a machine learning classifier that can detect programming languages, even for code snippets from a family of programming languages such as C, C++, and C\#. 
Khasnabish et al. \cite{khasnabishDetectingProgrammingLanguage2014} utilized several variants of Bayesian classifier models to detect 10 programming languages and achieved remarkable results. Multiple layers of neural networks and convolutional neural networks were trained to judge the programming language of over 60 kinds of source code \cite{gildaSourceCodeClassification2017b}. 
Large language models have demonstrated their enormous power in natural language tasks. Inspired by this, some researchers \cite{yangDeepSCCSourceCode2021a} have also employed large language models, such as RoBERTa, for source code classification with successful results.  {The RoBERTa classification method employs a pipeline that consists of RoBERTa Model and fully connected neural networks.}
{
	The CodeBERT model has been extensively utilized for code language classification. In the work of Liu et al. \cite{liuELCodeBertBetterExploiting2022}, a two-step fine-tuning method EL-CodeBERT was devised to address code smell detection. 
	This method utilizes the CodeBERT model to generate multiple layers of '[CLS]' vectors, which function as semantic representations for an input comment of source code. Subsequently, a bidirectional long short-term memory (LSTM) network is utilized to extract more effective features from these multiple layers of '[CLS]' vectors. In order to capture the significance of these features, an attention mechanism is employed to aggregate the most important ones. Finally, the aggregated features are fed into a fully connected neural network to perform the classification task. Two-step fine-tuning is utilized to address the difference in model training between CodeBert and Bi-LSTM\cite{liuELCodeBertBetterExploiting2022}.
}

\subsubsection*{Code Comment classification}

Code comments are a powerful tool to help programmers understand and maintain code snippets, but different comments can have different intentions \cite{shinyamaAnalyzingCodeComments2018}.
Rabi et al. \cite{raniHowIdentifyClass2021} developed a multilingual approach to code comment classification that utilizes natural language processing and text analysis to classify common types of class comment information with high accuracy for Python, Java, and Smalltalk programming languages. The Naive Bayes classifier, the J48 tree model, and the Random Forest model underlie the classifier used.
To reveal the relationship between a code block and the associated comment's category, Chen et al. \cite{chenWhyMyCode2021} classified comments into six intention categories and manually labeled 20,000 code-comment pairs. These categories include "what", "why", "how-to-use," "how-it-is-done", "property", and "others".
{
PLM based methods, such as EL-CodeBERT\cite{liuELCodeBertBetterExploiting2022}, have demonstrated their effectiveness in code comment classification, yielding promising results.
}

\subsubsection*{Technical Debt Classificationn}
Delivering high-quality, bug-free software is the goal of all software projects. When programmers are limited by time or other resources, the code they deliver is either incomplete, requires rework, produces errors, or is a temporary workaround. Such incomplete or temporary workarounds are commonly referred to as ``technical debt" \cite{potdarExploratoryStudySelfAdmitted2014,brownManagingTechnicalDebt2010,wehaibiExaminingImpactSelfAdmitted2016} at the cost of paying a higher price later on. Self-admitted technical debt (SATD) is common in software projects and can have a negative impact on software maintenance. Therefore, identifying SATD is very important for software engineering and needs to be investigated.
Potdar and Shihab \cite{potdarExploratoryStudySelfAdmitted2014} identified SATD by studying source code comments of four projects and manually devising an approach of 62 patterns. 
As manually designed patterns have significant drawbacks such as less generality and some physical burden, Huang et al. \cite{huangIdentifyingSelfadmittedTechnical2018a} investigated a text mining based approach that combines multiple classifiers to detect SATD in source comments of target software projects.
Since various characteristics of SATD features in code comments, such as vocabulary diversity, project uniqueness, length, and semantic variations, pose a notable challenge to the accuracy of pattern or traditional text mining-based SATD detection, Ren et al. \cite{renNeuralNetworkbasedDetection2019}  propose a convolutional neural network (CNN)-based approach for classifying code comments as SATD or non-SATD.
To avoid the daunting manual effort of extracting features, Wang et al. \cite{wangDetectingExplainingSelfadmitted2021} leverage attention-based neural networks to detect SATD.
{
EL-CodeBERT\cite{liuELCodeBertBetterExploiting2022}, a novel method that leverages a PLM, showcases its potential in SATD detection tasks.
}

\subsubsection*{Code Smell Classification}

Kent Beck defined the term ``code smell'' in the context of identifying quality problems in code that can be refactored to improve the maintainability of software \cite{fowler2018refactoring}, caused by design flaws or bad programmer habits.
Previous works have investigated various methods for identifying code smells in source code, and among them, machine learning is an effective approach for code smell classification. 
Fontana et al. \cite{arcellifontanaCodeSmellSeverity2017,arcellifontanaComparingExperimentingMachine2016} studied 16 different machine learning methods on four code smells (Data Class, Large Class, Feature Envy, Long Method) and 74 software systems with 1986 manually validated code smell samples. They found that J48 and Random Forest are able to achieve high performance in code smell classification.
Das et al. \cite{dasDetectingCodeSmells2019} propose a supervised convolutional neural network-based approach to code smell classification by eliminating the effort of manually selecting features.
To eliminate the manual effort of feature extraction, several neural network based methods have been proposed in the community. Neural network based methods still require a large number of labeled samples, and Liu et al. \cite{liuDeepLearningBased2021} proposed an automatic approach to generate labeled training data for neural network based classifiers.
Transfer learning can be a reliable solution to the dilemma of better performance and more labeled samples. Sharma et al. \cite{sharmaCodeSmellDetection2021}  propose a transfer learning method involving convolutional neural networks and recurrent neural networks, which can transfer the learned detection ability between C\# and Java language.
Li et al. \cite{liMultiLabelCodeSmell2022} propose a hybrid model based on deep learning for multi-label code smell classification, which utilizes both graph convolutional neural networks and bidirectional long short-term memory networks with attention mechanism.
{To capitalize on the benefits offered by large language models and enhance performance, EL-CodeBERT \cite{liuELCodeBertBetterExploiting2022} is employed for code smell classification. }

\subsection*{Prompt Learning Method}

Prompt-learning method has emerged as the fourth paradigm \cite{liuPretrainPromptPredict2021} of natural language processing (NLP)  and has drawn enormous attention from the NLP community, which adapts a variety of downstream NLP tasks to pre-trained tasks of large language models. Starting from GPT-3 \cite{brownLanguageModelsAre2020}, prompt learning has demonstrated its unique advantage in various downstream NLP tasks, with applications in text classification \cite{sunHowFineTuneBERT2019}, machine translation \cite{radford2019language}, etc. Existing researches focusing on prompt learning consists of three main components: a pre-traind language model, a prompt template and a verbalizer, in which  research related to verbalizer is orthogonal to our study. Previous studies related to PLM have broadly focused on PLM architectures, including BERT \cite{devlinBERTPretrainingDeep2019}, RoBERTa  \cite{liuRoBERTaRobustlyOptimized2019}, GPT-3, Bart \cite{lewisBARTDenoisingSequencetoSequence2020}, and others. Recently, several PLM related studies have focused on the software engineering domain, with CodeBERT \cite{kwonCodeBERTBasedSoftware2023} and GraphCodeBERT  \cite{guoGraphCodeBERTPretrainingCode2021} being typical works. As a generative PLM, CodeT5 \cite{wangCodeT5IdentifierawareUnified2021}  benefits a broad set of source code related tasks. A prompt template is used as a container to wrap the input text sequence into a prompt, which is then fed into a PLM to motivate the PLM to recall the rich knowledge associated with the input information. Templates can be constructed in a handcrafted manner \cite{perezTrueFewShotLearning2021,schickFewShotTextGeneration2021b} and achieve remarkable performance on a variety of downstream NLP tasks. To avoid the onerous effort of manually constructing a prompt template, some researchers seek to construct prompt templates automatically.
The automatically generated templates are of two types: discrete templates and continuous templates. The MINE approach \cite{jiangHowCanWe2020} is a mining-based discrete approach to automatically find templates given a set of training inputs x and outputs y.  Jiang et al. \cite{jiangHowCanWe2020} leverage round-trip translation of the prompt into another language then back to generate new templates. The MINE approach is a mining-based approach to automatically find templates given a set of training inputs x and outputs y. Prefix Tuning \cite{liPrefixTuningOptimizingContinuous2021} is a method that can be applied to continuous templates. The technique involves adding a sequence of task-specific vectors as prefixes to the input while keeping the parameters of the pre-trained language model (PLM) frozen.


{EL-CodeBERT establishes itself as the strongest baseline by attaining state-of-the-art performance across all four  source code-related tasks.}

\subsection*{Novelty of Our Study}

{
The proposed method represents a significant departure from existing state-of-the-art approaches that primarily rely on the '[CLS]' vectors obtained from multiple layers of the output vectors from a BERT-based model to represent input text (Liu et al., 2022; Jawahar et al., 2019), coupled with the use of LSTM to enhance features. In contrast, CodeClassPrompt introduces a novel approach that leverages prompt learning to induce knowledge features from a large language model. This innovative technique enables the successful completion of source code-related tasks without additional LSTM layers while reducing computational overhead during execution. Notably, CodeClassPrompt avoids a two-step training procedure by eliminating the need for LSTM layers.
Specifically, CodeClassPrompt utilizes prompt learning to guide a pre-trained language model towards generating outputs that are abundant in knowledge and intricately connected to a provided input sequence. To effectively capture indispensable feature information while discarding extraneous details, attention values are meticulously computed for a designated task using a specialized attention mechanism. Subsequently, the important features are aggregated and categorized without necessitating supplementary layers for feature extraction. This approach, employed by CodeClassPrompt, facilitates the extraction of comprehensive feature information while concurrently minimizing superfluous complexity.
}





\begin{figure*}
	\includegraphics[width=1.0\textwidth]{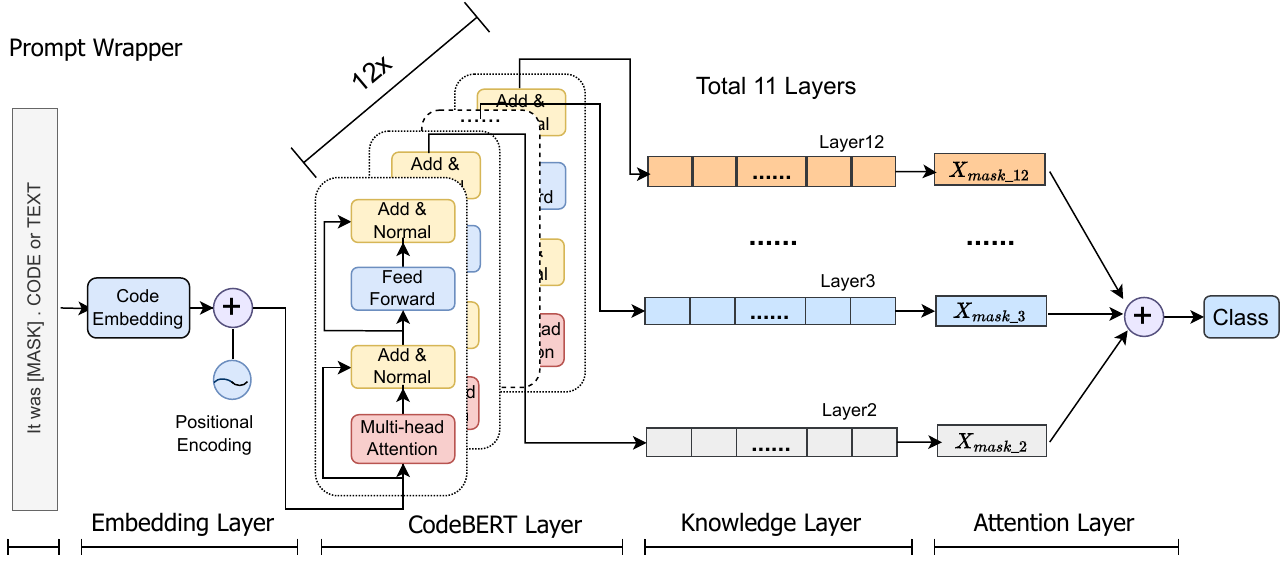}
	\caption{The architecture of CodeClassPrompt}
	\label{fig:1}       
\end{figure*}

\section*{CodeClassPrompt}

Our proposed approach, named CodeClassPrompt, leverages multiple aspects of knowledge recalled by a PLM to facilitate source code-related classification tasks. {The knowledge contained in each layer of CodeBERT output vectors has its own hierarchy \cite{jawaharWhatDoesBERT2019} and is regarded as a distinct aspect of knowledge.} The model architecture of our CodeClassPrompt approach is shown in Fig. \ref{fig:1}, which consists of a prompt wrapper, an embedding layer, a CodeBERT layer, a knowledge layer, and an attention layer.
\subsection*{{Task Definition}}
{
The CodeClassPrompt approach leverages a prompt template, which encapsulates segments of source code as a prompt, to activate a pre-trained model aimed at retrieving relevant knowledge associated with the input content. The acquired knowledge serves as the representation of the input information, appearing at the position of '[MASK]' in each layer of the output vector. Subsequently, an attention mechanism is employed to aggregate multiple layers of this representation, resulting in a feature representation of the input text. This aggregated feature is then utilized for the classification of the source code or related text.
}

{Specifically, we denote the pre-trained model as $M$, with CodeBERT being an example. We use $\mathbf{x}_p$ to represent the prompt template, where "$It\quad was\quad [MASK]$" serves as the prefix of the prompt, $\mathbf{x}$ denotes  the input text.}
\begin{equation*}
	\mathbf{x}_p= \texttt{[CLS]} It\quad was \quad\texttt{[MASK]} .\quad \mathbf{x}\nonumber 
\end{equation*}

 {
Prompt Learning harnesses the inherent capability of masked language modeling in BERT family models. This approach entails predicting pertinent contextual knowledge for masked positions based on diverse prompt prefixes associated with the original input text.
}

{
Let the input information $\mathbf{x}="int\quad main()\quad\{return \quad0;\}"$. After encapsulating $\mathbf{x}$ within the prompt template $\mathbf{x}_p$, During this process, $\mathbf{x}$ undergoes a transformation and gives rise to a modified representation denoted as $\mathbf{\hat{x}}_p$, which is commonly referred to as a prompt.
}

\begin{equation*}
	\mathbf{\hat{x}}_p= \texttt{[CLS]} It\quad was \quad\texttt{[MASK]} .\quad int\quad main()\quad\{return \quad0;\}\nonumber 
\end{equation*}

{
The aforementioned procedure is recognized as the Prompt Wrapper in this study. Following this, the modified representation $\mathbf{\hat{x}}_p$ is introduced as the input content that is fed into the model $M$. The CodeBERT model consists of two primary components: the embedding layer and the CodeBERT Layer. The embedding layer is responsible for mapping the input prompt $\mathbf{\hat{x}}_p$ to a set of vectors, each vector having a dimension size of 768. These vectors traverse the CodeBERT layer, which retrieves the relevant knowledge information stored during the pre-training phase of the CodeBERT model. This information is represented as a vector at the position of the '[MASK]' token within the output vectors. The overall output consists of 13 layers of vectors. The first layer corresponds to the embedded vectors of the input, while the subsequent layers capture hierarchical knowledge that is closely associated with the input content\cite{jawaharWhatDoesBERT2019}.
}

{
	The knowledge layer assumes the role of selecting multiple layers of vectors as input vectors for the attention layer, specifically targeting vectors located at the position of "[MASK]" within each layer. In the scope of this study, the selected layers encompass layer 2 through layer 12.
	}

The attention layer assigns individual weights to the different layers of the output derived from the knowledge layer. The weighted sum of these layer outputs serves as the representation that is then fed into the final classifier, which is a fully connected neural network. {The purpose of this process is to obtain the ultimate class label for the original input text.}

\subsection*{Prompt Wrapper}
The prompt wrapper is designed to wrap an input sequence into a prompt template as a prompt, which is then inputted into a PLM. A effective prompt can  induce the PLM to output relevant knowledge related to the input sequence.

In prompt learning, an input sequence $\mathbf{x}=\{w_1,w_2,...w_n\}$ need to first be wrapped into a prompt template 
\begin{equation}
	\mathbf{x}_{p}=\texttt{[CLS]} It\quad was \quad\texttt{[MASK]} .\quad \mathbf{x}\nonumber
\end{equation} as a prompt $\mathbf{\hat{x}}_p$,
then fed the prompt into a PLM.
\subsection*{Embedding Layer}
This layer is aimed to capture the relationship between tokens which maps target input from a textual form to a vector representation on a low-dimentional dense space. 
To the input a prompt to the model, the wrapped prompt text sequence is first tokenized by word-piece algorithm then we obtain the sequence $\mathbf{x} =(x_1,x_2,...,x_n)$, where $n$ is the length of the input sequence, $x_n$ is the $n$-th  tokenized sub-word.
Since the length of the input sequence varies for different inputs, we need to pad them to a uniform length to facilitate subsequent processing of the CodeBERT model. 
For the set maximum length of input sequence $N$,  if the length of sequences is less than  $N$, we pad 0 to the end of these sequences to make their length equal to $N$. For sequences whose length is greater than $N$, we directly truncate redundant text sequence at the end.
Therefore, the output of the embedding layer is $\mathcal{X}=(X_1,X_2,...X_N)$.
An input text is not merely a combination of tokens, it has important order information. To make the model to leverage the order information, absolute positional encoding (APE) is added in $\mathcal{X}$, the final output to feed into the next layer is 
\begin{equation}
	\mathcal{X}=\mathcal{X}+ APE(\mathcal{X}),  \mathcal{X}  \in  \mathbb R^{N\times d_{model}} 
\end{equation}
, where $d_{model}$ is the size  of the embedding dimension.

\subsection*{CodeBERT Layer}

In this layer, unlike other works, we leverage the knowledge information output from a masked language model. The embedding vectors from one embedding layer are fed into the layer to motivate the layer to output the knowledge information stored during the pre-training phase. The motivated knowledge information is leveraged as knolwedge features of input sequences. CodeBERT is a bi-modal pre-trained language model based on transformers for both programming languages (PL) and natural langauges (NL) \cite{fengCodeBERTPreTrainedModel2020b}. 
CodeBERT is pre-trained on a large general-purpose corpus by two tasks: Masked Language Model (MLM) and Replaced Token Detection (RTD). Specifically, the MLM task targets bimodal data by simultaneously feeding the code with the corresponding comments and randomly selecting positions for masking, then replacing the token with a special '[MASK]' token, the goal of the MLM task is to predict the original token. The RTD task targets unimodal data with separate codes and comments, randomly replaces the token, and aims to learn whether the token is the original word using a discriminator \cite{fengCodeBERTPreTrainedModel2020b}. However, in a large language model,such as CodeBERT, the output of '[MASK]' location is  representing not only the original token but also the relavent knowledge about input information.


\subsection*{Knowledge Layer}

Previous studies leverage only the vector of one or more layers at '[CLS]' location \cite{liuELCodeBertBetterExploiting2022}.
According to Choi et al. \cite{choiEvaluationBERTALBERT2021}, the output of '[CLS]' location is not the best choice for the representation of an input sequence, which are not stable for downstream tasks. Jawahar et al. \cite{jawaharWhatDoesBERT2019} proposed that each layer of the output of a BERT-like model has different semantic features that can be combined to further extract higher-level features that better represent the input. Unlike these studies, we leverage knowledge information from the output of CodeBERT.
As described in the prior subsection, the output of '[MASK]' location includes knowledge information related to the input, in which each layer has different aspect of knowledge and different importance related to the input. 
The output of the CodeBERT model has 13 layers, where one embedding layer and 12 layers for the encoders. The lowest-level embedding has little knowledge from the whole input, and the remaining layers have different importance. Through pilot experiments, the outputs from layer 2 through layer 12 have been 
chosen as knowledge sources.

\begin{equation}
	\mathcal{X}_{know}= \left[ 
	\begin{matrix}
		\mathbf{x}_{mask\_2} \\ \mathbf{x}_{mask\_3} \\ ... \\ \mathbf{x}_{mask\_12} 
	\end{matrix} 
	\right], \mathcal{X}_{know} \in \mathbb{R}^{ 11\times d_{model}}
\end{equation}

, where $d_{model}$ is the size  of the embedding dimension.

\subsection*{Attention Layer}
Several knowledge features were obtained in the previous subsection, not all representational information contributes equally to the input, and each layer of knowledge features has different weights for the entire representation of the input.  Some source code related tasks focus more on lower-level knowledge features, while others focus more on higher-level features. Therefore, in CodeClassPrompt, the attention mechanism is used to compute different weights for each knowledge feature.
Specifically, we first compute the tanh value of $\mathcal{X}_{know}$ as

\begin{equation}
	\mathbf{u}_i = tanh(\mathcal{X}_{know})
\end{equation}
, then  the similarity between $\mathbf{u}_i$ and the context vector $\mathbf{u}_w$ can be calculated and transformed into a probability distribution by Softmax.

\begin{equation}
	\mathbf{\alpha}_i=\frac{\exp{(\mathbf{u}_i^T \mathbf{w}_u)}}{\sum_{i}\mathbf{u}_i^T \mathbf{w}_u}
\end{equation}

$\mathbf{\alpha}_i$ can be treated as the importance of the input for each leavel of knowledge feature, therefore using $\mathbf{\alpha}_i$ as a global weighted summation over $\mathcal{X}_{know}$ can generate the input vector $\mathbf{x}_{out}$,

\begin{equation}
	\mathbf{x}_{out}=\sum_{i} \mathbf{\alpha}_i\mathbf{x}_i
\end{equation}

Finally, for the $\mathbf{x}_{out}$, it can be classified by a layer of fully connected feed-forward network.

\begin{equation}
	\mathbf{p}(y|\mathbf{x}_{out})=\mathbf{w}( ReLU(\mathbf{x}_{out}))+b
\end{equation}

The final class label $y$ of the input is
\begin{equation}
	\hat{y}=\arg\max{\mathbf{p}(y|\mathbf{x}_{out})}
\end{equation}

\section*{Experimental Study}

In this section, we design four source code-related tasks aimed at answering the following questions (RQs).

\textbf{RQ1}: \quad {Can our proposed approach achieve comparable results for source code-related tasks without the necessity of additional feature extraction layers, in comparison to the baselines?}

\textbf{RQ2}:  \quad Can the performance of CodeClassPrompt be enhanced by solely utilizing either the attention mechanism or a prompt template?

\textbf{RQ3}: \quad How does the attention mechanism work on the four source code-related classification tasks?

\textbf{RQ4}: \quad Does the proposed model perform equally well on both programming language-based and natural language-based tasks?

\textbf{RQ5}: \quad { What is the potential improvement in computational efficiency that can be achieved by eliminating additional neural network layers?}

RQ1 aims to validate the performance of CodeClassPrompt on four source code-related classification tasks. To achieve this goal, we conducted extensive experiments and compared CodeClassPrompt with current state-of-the-art baselines. In RQ2, we investigated the effectiveness of different components in the CodeBERT-based approach on performance. To answer these questions, we performed comprehensive ablation studies on four datasets. RQ3 examines the differences in focusing on different layers of knowledge across four different tasks. RQ4 addresses the question of whether knowledge features are equally effective for both natural language-based and programming language-based tasks. Lastly, {RQ5 explores the potential improvement in computational efficiency that can be attained by eliminating additional neural network layers within a CodeBERT pipeline.}

\subsection*{Tasks and Datasets}
To validate our proposed approach, we conducted extensive experiments on four downstream tasks related to source code. The first two tasks (code language classification and code smell classification) are related to programming language processing, while the other two (code comment classification and technical debt classification) are related to natural language processing. Each task has a dedicated dataset, which is described in detail in the corresponding task description.

\begin{table}[!ht]
	\caption{Corpus statistics for four source code-related tasks}
	\label{tab:datasets}
	\footnotesize  
	\resizebox*{\linewidth}{!}{%
		\begin{tabular}{@{}llcccccccccc@{}}
			\toprule
			Task & Class Num& Train & Test &Avg &Mode & Median &  \textless 32 &  \textless 64 &  \textless 128 & \textless 256 & \textless 300 \\ 
			\midrule
			
			Code Language &19 & 179,556&44,889& 58.58\%&2& 31.0& 51.04\%& 73.76\%& 89.78\%&96.94\%&97.75\% \\ [0.2cm]
			Code Smell & 2 & 1,399 & 350 & 41.97\%&8 & 20.0& 68.19\%& 85.56\%& 93.14\%& 97.63\%&98.43\%\\ [0.2cm]
			Code Comment &16& 8,985&2,247&15.82\%&2&7.0&88.46\%&93.88\%&97.12\%&99.97\%&99.99\% \\ [0.2cm]
			Technical Debt &2&31,708 &6,652 &9.27\%&3& 6.0& 95.98\%& 99.23\%& 99.87\%&99.98\%&99.99\% \\ [0.2cm]
			
			\bottomrule
		\end{tabular}%
	}
	
\end{table}

	\begin{table}[!ht]
	\centering
	\caption{The statistical information of dataset CODE LANGUAGE: train set.}
	\label{tab:dataset_CODE_train}
	\begin{tabular}{ll|ll|ll|ll|ll}
		\toprule
		Label & Count & Label & Count & Label & Count& Label & Count & Label & Count  \\
		\midrule
		0 &     9574 & 1 &      9605 & 	2 &   9594 &  3 &     9559 & 4 &      9639 \\
		\midrule
		
		5 & 9681  &	6  &    9584&	7  & 9542 &	8  &      6780 	 &		9  & 9591	\\
		\midrule
		10 &     9623 &	11 &     9546&	12 &     9556&	13&     9639&	14 &     9611 									\\
		\midrule
		15 &     9660 &	16 &   9591&	17 &    9527&	18 &     9654	&	 &		\\
		\bottomrule
	\end{tabular}
\end{table}

		\begin{table}[!ht]
	\centering
	\caption{The statistical information of dataset CODE LANGUAGE: test set.}
	\label{tab:dataset_CODE_test}
	\begin{tabular}{ll|ll|ll|ll|ll}
		\toprule
		Label & Count & Label & Count & Label & Count& Label & Count & Label & Count  \\
		\midrule
		0 &   2427 	&	1 &      2396&	2&      2407 &		3 &      2442&		4 &      2362 \\
		\midrule
		
		5 &  2320		&	6 &     2417	&		7&      2459	&		8&      1647&			9&     2410	\\
		\midrule
		10 &    2378 	&	11 &     2455	&	12&     2445	&	13 &     2362	&	14 &     2390									\\
		\midrule
		15 &    2341 	&	16 &     2410	&	17 &     2474	&	18&    2347	&	 &		\\
		\bottomrule
	\end{tabular}
\end{table}

\begin{table}[!ht]
	\centering
	\caption{The statistical information of dataset CODE SMELL: train set.}
	\label{tab:dataset_SMELL_train}
	\begin{tabular}{ll|ll}
		\toprule
		Label & Count & Label & Count  \\
		\midrule
		0 &   755 	&	1 &      644 \\
		
		\bottomrule
	\end{tabular}
\end{table}

	\begin{table}[!ht]
	\centering
	\caption{The statistical information of dataset CODE SMELL: test set.}
	\label{tab:dataset_SMELL_test}
	\begin{tabular}{ll|ll}
		\toprule
		Label & Count & Label & Count  \\
		\midrule
		0 &   189 	&	1 &      161 \\
		
		\bottomrule
	\end{tabular}
\end{table}

\begin{table}[!ht]
	\centering
	\caption{The statistical information of dataset CODE COMMENT: train set.}
	\label{tab:dataset_COMMENT_train}
	\begin{tabular}{ll|ll|ll|ll|ll}
		\toprule
		Label & Count & Label & Count & Label & Count& Label & Count & Label & Count  \\
		\midrule
		0 &     164 &			1 &     263		&	2 &     43	&		3&      782		&	4 &     195 \\
		\midrule
		
		5 &      159	&	6 &      46	&	7 &     70	&	8 &     662	&	9&      357	\\
		\midrule
		10&     823 	&		11 &     205	&		12 &     3365&			13 &     152	&		14 &     16								\\
		\midrule
		15 &     1683	&	 &   	&	 &   	&	&   	&	 &		\\
		\bottomrule
	\end{tabular}
\end{table}

\begin{table}[!ht]
	\centering
	\caption{The statistical information of dataset CODE COMMENT: test set.}
	\label{tab:dataset_COMMENT_test}
	\begin{tabular}{ll|ll|ll|ll|ll}
		\toprule
		Label & Count & Label & Count & Label & Count& Label & Count & Label & Count  \\
		\midrule
		0 &     41		&	1 &      66		&	2 &     11	&		3 &     196		&	4 &     49 \\
		\midrule
		
		5 &     40	&	6 &      11	&	7 &      17	&	8 &     166	&	9 &      89	\\
		\midrule
		10 &     206	&		11 &     51		&	12 &    841	&		13 &     38		&	14 &     4								\\
		\midrule
		15 	&     421 	&	 &     	&	&     	&	&   	&	 &		\\
		\bottomrule
	\end{tabular}
\end{table}

	\begin{table}[!ht]
	\centering
	\caption{The statistical information of dataset CODE SATD: train set.}
	\label{tab:dataset_SATD_train}
	\begin{tabular}{ll|ll}
		\toprule
		Label & Count & Label & Count  \\
		\midrule
		0 &   29020 	&	1 &      2688 \\
		
		\bottomrule
	\end{tabular}
\end{table}

\begin{table}[!ht]
	\centering
	\caption{The statistical information of dataset CODE SATD: test set.}
	\label{tab:dataset_SATD_test}
	\begin{tabular}{ll|ll}
		\toprule
		Label & Count & Label & Count  \\
		\midrule
		0 &   6069 	&	1 &      583 \\
		
		\bottomrule
	\end{tabular}
\end{table}

\subsubsection*{Code Language Classification}

In this task, the target data are source code snippets. We leverage the publicly shared dataset in SC++ \cite{alrashedySCCPredictingProgramming2020a}, which collects 21 programming languages popular in the Stack Overflow community, based on the 2017 Stack Overflow developer survey  \footnote{https://insights.stackoverflow.com/survey/2017\#technology}.  As is inevitable, there is some invalid source code in the dataset, which we removed using tools from DeepSCC \cite{yangDeepSCCSourceCode2021a}, keeping 19 programming languages. In other words, this task is a multi-class source code task that predicts the programming language type of each code snippet.

\subsubsection*{Code Smell Classification}

Code smell classification is a binary classification task where a model needs to decide whether it has code smell or not, and its target is the source of a variety of code snippets. Fakhoury et al. \cite{fakhouryKeepItSimple2018} build a corpus which selects 4205 lines of source code from 13 java open-source systems to avoid domain-specific dependencies of the results. The corpus has over 1700 labeled code snippets, which is in line with the taxonomy of linguistic smells presented in the paper  \cite{arnaoudovaLinguisticAntipatternsWhat2016}. 
The corpus is a typical dataset for code smell classification in previous studies \cite{liuELCodeBertBetterExploiting2022}, and we adopt it for this task.

\subsubsection*{Code Comment Classification}
Code comments are crucial software components that contain important information concerning software design, code implementation, and other technical details. 
We adopt the corpus shared by Pascarella and Bacchelli \cite{pascarellaClassifyingCodeComments2017}, which has over 11, 000 code reviews and 16 classes from six java open source software projects. This task is a multi-class natural language classification task that aims to categorize each comment into a specific class.

\subsubsection*{Technical Debt Classification}

Techincal debt classification is a natural language classification task based on annotated information indicating whether there is technical debt or not. Our goal is to detect technical debt annotated by programmers, i.e., self-admitted techical debt (SATD). The dataset presented by Maldonado et al. \cite{maldonadoUsingNaturalLanguage2017}  consists of approximately 10, 000 code comments collected from 10 open source projects, which are classified into five types of SATD, namely, design debt, requirement debt, defect debt, documentation debt, or test debt. All our SATD experiments are performed on the corpus.

For brevity, in the following sections, code language classification is abbreviated as code language, code smell classification as code smell, code comment classification as code comment, and technical debt classification as technical debt.

{To ensure a rigorous and unbiased evaluation, we employ a stratified sampling technique to partition the corpus into distinct training and test sets for each task, adhering to an 80:20 percent ratio. This approach enables a fair comparison with the baseline results.}
These statistics consist of (1) the number of training and test sets, (2) the mean, mode, and median of the code/comment length, and (3) the percentage of samples with sizes \textless 32, \textless 64, \textless 128, \textless 256, and \textless 300.
All statistical information for the four datasets is detailed in Table 1.

{Table \ref{tab:dataset_CODE_train}, Table \ref{tab:dataset_CODE_test}, Table \ref{tab:dataset_SMELL_train}, Table \ref{tab:dataset_SMELL_test},  Table \ref{tab:dataset_COMMENT_train}, Table \ref{tab:dataset_COMMENT_test},   Table \ref{tab:dataset_SATD_train}, and Table \ref{tab:dataset_SATD_test} provide a comprehensive breakdown of the category distribution for each dataset. The Code Language and Code Smell datasets demonstrate a well-balanced distribution of classes without any significant class imbalance. In the case of the remaining datasets, some degree of class imbalance is observed, with certain classes being underrepresented compared to others.
}

\subsection*{Evaluation Metrics}

Accuracy, Precision, Recall, and F1-Score are chosen as evaluation metrics for the binary classification task. These evaluation metrics are calculated as follows.

\begin{equation}
	Accuracy= \frac{TP + TN}{TP+TN+FP+FN}
\end{equation}
\begin{equation}
	Precision = \frac{TP}{TP+FP}
\end{equation}
\begin{equation}
	Recall = \frac{TP}{TP+FN}
\end{equation}	

\begin{equation}
	F1\mbox{-}Score=\frac{2 \times Precision \times Recall}{Precision+Recall}
\end{equation}
where $TP$ means that a positive sample is predicted as a positive class, $TN$ means that a negative sample is predicted as a negative class, $FP$ means that a negative sample is assigned a positive label, and $FN$ means that a positive sample is predicted as a negative class.

For all multi-classification tasks, we leverage the macro approach to compute evaluation metrics. Specifically, we tally $TP$, $FP$, $FN$ and $TN$ for each class and then compute Precision, Recall and F1-Score, respectively. Finally, we obtain the mean value of each metric  for  all classes to obtain Macro-Precision, Macro-Recall, and Macro-F1-Score.

For simplicity, Accuracy is abbreviated as ACC, Precision as P, Recall as R, and F1-Score as F1 in the following tables.

\subsection*{Baselines}

Previous studies have extensively investigated source code-related tasks \cite{sharmaSurveyMachineLearning2022}, utilizing a range of AI-based tools from machine learning to deep learning.
{In this study, we conduct a comprehensive evaluation of our CodeClassPrompt model in comparison to eight recently proposed baselines across four source code-related classification tasks. These baselines can be categorized into two distinct groups based on the AI development perspective: machine learning-based approaches and neural network-based approaches. The latter category can be further divided into two sub-classes, namely classical neural network-based approaches and pre-trained language model-based methods.}
The following is a brief introduction to all the baselines we compare with: Random Forest, XGBoost, TextCNN, AttBLSTM, BERT, RoBERTa, CodeBERT, and EL-CodeBERT.

\subsubsection*{Machine Learning based Methods}

\textbf{Random Forest} is an ensemble machine learning algorithm based on decision trees and bagging proposed by Breiman \cite{breimanRandomForests2001}, and the baseline experiments were implemented with the library scikit-learn\footnote{https://github.com/scikit- learn/scikit-learn}.

\noindent\textbf{XGBoost} is sugguest by Chen and Guestrin \cite{chenXGBoostScalableTree2016a}, which is a scalable end-to-end tree boosting system and is used widely by data scientists to achieve state-of-the-art results on many machine learning challenges. It is sparsity-aware algorithm for sparse data and weighted quantile sketch for approximate tree learning. We adopt the offiical implementation from the original author on github\footnote{https://github.com/dmlc/xgboost}.

\subsubsection*{Neural Networks based Methods}

\noindent\textbf{Classical Neural Netowrk based Approaches}

\noindent\textbf{TextCNN} \quad Kim \cite{kimConvolutionalNeuralNetworks2014b} proposed TextCNN, a sophisticated approach that leverages convolutional neural networks for natural language processing. In this study, we adopt a classical implementation of TextCNN as a baseline\footnote{https://github.com/NTDXYG/Text-Classify-based-pytorch/blob/master/model/TextCNN.py}.

\noindent\textbf{AttBLSTM}\quad Zhou et al. \cite{zhouAttentionBasedBidirectionalLong2016} proposed AttBLSTM, a combined approach that leverages attention mechanism and bidirectional long short-term memory network (BiLSTM) to capture important semantic features of textual sequences. In this study, we utilize a baseline implementation of AttBLSTM based on source code available on Github\footnote{https://github.com/NTDXYG/Text-Classify-based-pytorch/blob/master/model/TextRNN\_Attention.py}.

\noindent\textbf{Pre-trained Model based Methods}

\noindent\textbf{BERT} \cite{devlinBERTPretrainingDeep2019} is a deep bidirectional transformer-based language model for language understanding, pre-trained on both next sentence prediction and masked language modeling tasks using self-supervised methods with large-scale corpora. In this study, we utilize a baseline implementation of BERT based on the BERT-base model\footnote{https://huggingface.co/transformers/v3.0.2/model\_doc/bert.html\#bertforsequenceclassification}.

\noindent\textbf{RoBERTa} \cite{liuRoBERTaRobustlyOptimized2019} is an enhanced BERT-based model pre-trained on a much larger corpus than the original BERT model\footnote{https://huggingface.co/roberta-base} using only masked language modeling approach. In this study, we utilize a baseline implementation of sequence classification based on the RoBERTa-base model, which is the official implementation provided by the authors\footnote{https://huggingface.co/docs/transformers/model\_doc/roberta\#transformers.RobertaForSequenceClassification}.

\noindent\textbf{CodeBERT} \cite{fengCodeBERTPreTrainedModel2020b} is a transformer-based model pre-trained on a corpus consisting of both programming language and natural language, using both masked language modeling and replaced token detection tasks. In this study, we utilize a baseline implementation of CodeBERT based on the CodeBERT-base model.

\noindent\textbf{EL-CodeBERT} \cite{liuELCodeBertBetterExploiting2022} is a two-stage model that builds upon CodeBERT, attention mechanism, and BiLSTM. The model utilizes BiLSTM to extract multiple layers of semantic features and then employs attention mechanism to aggregate the final features, resulting in promising performance compared to other baselines. To ensure consistency with the baseline implementation, we adopt the official code and training method\footnote{https://github.com/NTDXYG/EL\-CodeBert}.

\subsection*{Experimental Settings}

\begin{table}[!ht]
	\centering
	\caption{Prompt templates for four tasks}
	\label{tab:prompt}
	\resizebox{0.5\textwidth}{!}{%
		\begin{tabular}{@{}ll@{}}
			\toprule
			Task & Prompt Template  \\ 
			\midrule
			
			{ Code Language } & " Just [MASK] ! \textbf{x}"\\
			{ Code Smell } &  " \textbf{x} In summary , it was [MASK] ."\\
			{ Code Comment } &  " It was [MASK] . \textbf{x}"\\
			{ Technical Debt } &  " Just [MASK] ! \textbf{x}"\\

			\bottomrule
		\end{tabular}%
	}
\end{table}
Our proposed approach utilizes OpenPrompt \cite{ding2021openprompt}, an open-source framework for prompt learning, along with the CodeBERT-base model \cite{fengCodeBERTPreTrainedModel2020b}.To ensure consistent and fair comparison, we conducted all experiments using an Nvidia RTX3090 GPU (Graphic Processing Unit), Linux operating system (Ubuntu 22.04), and 64GB of system memory for both the baselines and our proposed method.

\begin{table}[!ht]
	\centering
	\caption{All template candidates}
	\label{tab:prompt_all}
	\resizebox{0.4\textwidth}{!}{%
		\begin{tabular}{@{}ll@{}}
			\toprule
			No. & Prompt Template  \\ 
			\midrule
			
			1 & " It was [MASK] . \textbf{x}" \\
			2 &  " \textbf{x} In summary , it was [MASK] ."\\
			3 &  " \textbf{x} All in all , it was {"mask"} ."\\
			4 &  " Just [MASK] ! \textbf{x}"\\

			\bottomrule
		\end{tabular}%
	}
\end{table}

{
The prompt templates leveraged for each respective dataset are delineated in Table \ref{tab:prompt}, which were selected from the four universal template candidates enumerated in Table \ref{tab:prompt_all}, derived from the OpenPrompt Framework \cite{ding2021openprompt}.
To ascertain the most suitable template for each task, a series of pilot experiments were undertaken. These experiments served as a systematic evaluation to assess the performance and alignment of each template with the specific task requirements.
}

\section*{Results and Analysis}

This section presents the evaluation and analysis of our proposed approach, CodeClassPrompt. We will commence by comparing our results with the baselines on four source code-related tasks to demonstrate the performance of our approach. {All results of CodeClassPrompt are obtained by conducting five independent runs with different random seeds. The average value and standard deviation are calculated for each metric, with the maximum value indicated in parentheses.} Following that, we will experimentally validate the effectiveness of each component in our approach. We will then conduct a series of attention-related experiments to demonstrate how different tasks focus on different knowledge layers. Next, we will analyze the varying effects of knowledge features on programming language-based and natural language-based tasks. {Finally, a series of experiments will be conducted to demonstrate the computational cost-saving potential of CodeClassPrompt. }

\subsection*{Result Analysis for RQ1}

\begin{table}[!ht]
	\centering
	\caption{{Results on code language classification task.} The values presented in \textbf{Bold} denote the optimal values achieved for each metric.}
	\label{tab:codelang}
	\resizebox{0.95\textwidth}{!}{%
		\begin{tabular}{@{}lllll@{}}
			\toprule
			Method & ACC(\%) & P(\%) & R(\%) & F1(\%) \\ \midrule
			
			{ Random Forest\cite{liuELCodeBertBetterExploiting2022} } & 78.728 & 79.362 & 78.825 & 78.874 \\
			{ XGBoost\cite{liuELCodeBertBetterExploiting2022} } & 78.803 & 79.925 & 78.891 & 79.217 \\
			{ TextCNN\cite{liuELCodeBertBetterExploiting2022}*} & 82.662 & 83.561 & 82.706 & 82.964 \\
			{ AttBLSTM\cite{liuELCodeBertBetterExploiting2022}*} & 79.035 & 79.801 & 79.107 & 79.272 \\
			{ BERT\cite{liuELCodeBertBetterExploiting2022} } & 86.865 & 87.129 & 86.938 & 86.985 \\
			{ RoBERTa\cite{liuELCodeBertBetterExploiting2022} } & 87.202 & 87.424 & 87.276 & 87.135 \\
			{ CodeBERT\cite{liuELCodeBertBetterExploiting2022} } & 87.418 & 88.042 & 87.450 & 87.614 \\
			{ EL-CodeBERT(2022)\cite{liuELCodeBertBetterExploiting2022} } & {87.959} & {88.177} & {88.023} & {88.077} \\
			\midrule
			{ \textcolor{red}{EL-CodeBERT}} & {$87.757 \pm0.075(87.870)$} & {$88.065\pm0.058(88.139)$} & {$87.816\pm0.081(87.934)$} & {$87.895\pm0.073(88.002)$} \\
			
			{ CodeClassPrompt } & $87.906\pm0.085(\textbf{88.024})$ & $88.104\pm0.089(\textbf{88.232})$ & $87.980\pm 0.083(\textbf{88.091})$ & $88.030\pm0.085(\textbf{88.149})$ \\

			\bottomrule
		\end{tabular}%
	}
\end{table}

In Table \ref{tab:codelang}, \ref{tab:codesmell}, \ref{tab:codecomment}, and \ref{tab:codesatd}, the first two rows correspond to machine learning approaches, the next two rows marked with an asterisk (*)  are classical neural network approaches without pre-trained language models, and the next four rows are pre-trained model-based approaches. The number in parentheses following a method name denotes the year the method was proposed. { The method names of baselines highlighted in red signify that their results were obtained through experimentation, whereas the other baseline results are cited from the paper of EL-CodeBERT \cite{liuELCodeBertBetterExploiting2022}.}

\noindent \textbf{Code Language Classification.}\quad Table \ref{tab:codelang} presents the results of the comparison between CodeClassPrompt and all baselines.  We observe that pre-trained model-based methods achieve better performance than machine learning and classical neural networks, with clear advantages.

Our CodeClassPrompt approach demonstrates comparable performance to all baseline methods across the four evaluation metrics, namely accuracy, precision, recall, and F1-score. Specifically, the achieved values for accuracy, precision, recall, and F1-score are 87.906\% (with a maximum value of 88.024\%), 88.104\% (with a maximum value of 88.232\%), 87.980\% (with a maximum value of 88.091\%), and 88.030\% (with a maximum value of 88.149\%) respectively. Importantly, each maximum value surpasses the reported values of the baseline models. These findings serve as compelling evidence that the integration of knowledge features and attention mechanisms significantly enhances the performance of source code-related classification tasks. Notably, the utilization of knowledge features inspired by prompts has proven to be effective in capturing the salient features of the input information more accurately.

\begin{table}[!ht]
	\centering
	\caption{{Results on code smell classification task.} The values presented in \textbf{Bold} denote the optimal values achieved for each metric.}
	\label{tab:codesmell}
	\resizebox{0.95\textwidth}{!}{%
		\begin{tabular}{@{}lllll@{}}
			\toprule
			{ Method } &  { ACC }(\%) & {P}(\%) & {R}(\%) & {F1}(\%) \\
			\midrule
			{ Random Forest\cite{liuELCodeBertBetterExploiting2022} } & 78.286 & 78.880 & 77.548 & 77.756 \\
			{ XGBoost\cite{liuELCodeBertBetterExploiting2022} } & 75.714 & 75.667 & 75.305 & 75.409 \\
			{ TextCNN\cite{liuELCodeBertBetterExploiting2022}* } & 80.000 & 80.016 & 79.641 & 79.761 \\
			{ AttBLSTM\cite{liuELCodeBertBetterExploiting2022}*} & 78.857 & 78.810 & 78.537 & 78.631 \\
			{ BER\cite{liuELCodeBertBetterExploiting2022}T } & 79.714 & 79.580 & 79.653 & 79.610 \\ 
			{ RoBERTa\cite{liuELCodeBertBetterExploiting2022} } & 81.143 & 81.014 & 81.067 & 81.038 \\ 
			{ CodeBERT\cite{liuELCodeBertBetterExploiting2022} } & 85.429 & 85.516 & 85.128 & 85.264 \\ 
			{ EL-CodeBERT(2022)\cite{liuELCodeBertBetterExploiting2022} } & 86.000 & {85.990} & {85.795} & {85.874} \\
			\midrule
			{ \textcolor{red}{EL-CodeBERT} } & $76.343\pm1.134(77.714)$ & $76.752\pm1.000(77.927)$ & $75.767\pm1.380(77.479)$ & $75.854\pm1.343(77.526)$ \\
			
			{ CodeClassPrompt } & $\textbf{86.171}\pm0.291(\textbf{86.571})$ & $\textbf{86.329}\pm0.348(\textbf{86.678})$ &$\textbf{85.862}\pm0.33(\textbf{86.462})$ & $\textbf{86.006}\pm0.3(\textbf{86.479})$ \\

			\bottomrule
		\end{tabular}%
	}
\end{table}

\noindent\textbf{Code Smell Classification.}\quad {The comparative results between CodeClassPrompt and all baseline methods are presented in Table \ref{tab:codesmell}. Among the evaluated methods, the pre-trained model-based approach exhibited the highest performance. However, our CodeClassPrompt method outperformed all other methods across all metrics, achieving the highest accuracy, precision, recall, and F1-score values of 86.171\%, 86.329\%, 85.862\%, and 86.006\% respectively. {The empirical results demonstrate that CodeClassPrompt is capable of achieving superior performance on the Code Smell dataset compared to the state-of-the-art baseline approaches, all while requiring significantly lower computational resources. } This can be attributed to the superior feature extraction capabilities of the CodeClassPrompt method in the domain of programming language processing.}

\begin{table}[!ht]
	\centering
	\caption{Results on code comment classification task. The values presented in \textbf{Bold} denote the optimal values achieved for each metric.}
	\label{tab:codecomment}
	\resizebox{0.95\textwidth}{!}{%
		\begin{tabular}{@{}lllll@{}}
			\toprule
			Method & ACC(\%) & P(\%) & R(\%) & F1(\%) \\ \midrule
			
			{ Random Forest\cite{liuELCodeBertBetterExploiting2022} } & 90.921 & 83.783 & 75.104 & 74.618 \\
			{ XGBoost\cite{liuELCodeBertBetterExploiting2022} } & 90.565 & 77.561 & 68.661 & 71.645 \\
			{ TextCNN\cite{liuELCodeBertBetterExploiting2022}*} & 91.945 & 87.541 & 78.596 & 80.977 \\
			{ AttBLSTM\cite{liuELCodeBertBetterExploiting2022}*} & 92.345 & 85.578 & 78.268 & 80.596 \\
			{ BERT\cite{liuELCodeBertBetterExploiting2022} } & 94.482 & 87.149 & 83.935 & 85.275 \\
			{ RoBERTa\cite{liuELCodeBertBetterExploiting2022} } & 94.393 & \textbf{90.525} & 86.121 & 86.875 \\
			{ CodeBERT\cite{liuELCodeBertBetterExploiting2022} } & 94.838 & 87.916 & 86.301 & 86.820 \\
			{ EL-CodeBERT(2022)\cite{liuELCodeBertBetterExploiting2022} } &  \textbf{95.238} & 89.395 & \textbf{87.280} & \textbf{87.977} \\
			\midrule
			{ \textcolor{red}{EL-CodeBERT}  } &  $94.811\pm0.170(94.971)$& $89.070\pm2.886(92.522)$ & $85.966\pm0.651(86.530)$ & $86.673\pm1.083(87.475)$\\
		
			{ CodeClassPrompt } & $95.220\pm 0.108(\textbf{95.416})$ & $89.193\pm0.648(89.691)$ & $86.938\pm0.287{(87.183)}$ & $87.770\pm0.249({87.942})$ \\

			\bottomrule
		\end{tabular}%
	}
\end{table}

\noindent\textbf{Code Comment Classification.}\quad {
Table \ref{tab:codecomment} presents the results obtained from the comparative analysis between CodeClassPrompt and all the baseline models. The table reveals that CodeClassPrompt has achieved results that are comparable to those of the baselines. The metrics used for evaluation, namely accuracy, precision, recall, and F1-score, indicate values of 95.220\%, 89.193\%, 86.938\%, and 87.770\% respectively. Moreover, CodeClassPrompt has achieved an impressive maximum accuracy value of 95.416\%. These findings serve as evidence of CodeClassPrompt's commendable performance in the realm of code comment classification, underscoring its promising capability in extracting features from natural language-based content. It is worth noting that all of these results were obtained with significantly lower computational resources when compared to EL-CodeBERT.
}

\begin{table}[!ht]
	\centering
	\caption{{Results on technical debt classification task. The values presented in \textbf{Bold} denote the optimal values achieved for each metric.}}
	\label{tab:codesatd}
	\resizebox{0.95\textwidth}{!}{%
		\begin{tabular}{@{}lllll@{}}
			\toprule
			Method & ACC(\%) & P(\%) & R(\%) & F1(\%) \\ \midrule

			{ Random Forest\cite{liuELCodeBertBetterExploiting2022} } & 97.278 & \textbf{95.080} & 87.019 & 90.564 \\
			{ XGBoost\cite{liuELCodeBertBetterExploiting2022} } & 97.294 & 93.901 & 88.516 & 90.991 \\
			{ TextCNN\cite{liuELCodeBertBetterExploiting2022}*} & 96.978 & 93.015 & 87.258 & 89.881 \\
			{ AttBLSTM\cite{liuELCodeBertBetterExploiting2022}*} & 97.114 & 93.979 & 87.177 & 90.227 \\
			{ BERT\cite{liuELCodeBertBetterExploiting2022} } & 96.783 & 91.229 & 88.004 & 89.534 \\
			{ RoBERTa\cite{liuELCodeBertBetterExploiting2022} } & 97.595 & 93.352 & 91.110 & 92.279 \\
			{ CodeBERT\cite{liuELCodeBertBetterExploiting2022} } & 97.835 & 94.197 & 91.991 & 93.059 \\
			{ EL-CodeBERT(2022)\cite{liuELCodeBertBetterExploiting2022}  } & {97.850} & 94.024 & {92.310} & {93.146} \\
			\midrule
			{ \textcolor{red}{EL-CodeBERT}  } & $\textbf{97.895}\pm0.016(\textbf{97.910})$ & $93.687\pm0.713({94.970})$ & $\textbf{93.110}\pm0.994(\textbf{94.273})$ & $\textbf{93.375}\pm0.175(\textbf{93.530} )$ \\
		
			{ CodeClassPrompt } & $97.811\pm0.058({97.895})$ & $93.954\pm 0.234(94.330)$ & $92.118\pm 0.248({92.387}) $& $93.011\pm0.186({93.263})$ \\
			
			\bottomrule
		\end{tabular}%
	}
\end{table}

\noindent\textbf{Technical Debt Classification.}\quad {
Table \ref{tab:codesatd} presents the outcomes of the comparative analysis conducted between CodeClassPrompt and all the baseline approaches. In terms of the evaluation metrics, {CodeClassPrompt has demonstrated comparable results}, achieving accuracy, precision, recall, and F1-score values of 97.811\%, 93.954\%, 92.118\%, and 93.011\% respectively. The table reveals that all the investigated approaches, ranging from machine learning methods to pre-trained model-based techniques, have yielded similar accuracy levels. 
}

\noindent\textbf{Summary for RQ1:} 
{
The results demonstrate that CodeClassPrompt exhibits a clear superiority in performance on the code smell classification task, while also achieving comparable levels of performance to the state-of-the-art baseline approaches across three additional evaluated tasks. These findings underscore the operational efficacy and inherent strengths of the CodeClassPrompt methodology. Specifically, this approach distinguishes itself in tasks associated with programming language processing, such as code language classification and code smell identification, offering unique benefits. Importantly, these findings emphasize that CodeClassPrompt not only matches the performance of previous state-of-the-art baselines but does so with reduced computational costs. This efficiency can be primarily attributed to CodeClassPrompt's superior capability in extracting features from both programming and natural languages, eliminating the need for additional neural network layers to enhance its feature extraction process.
}

\begin{table}[!ht]
	\centering
	\caption{{Results of the T-test conducted between CodeClassPrompt and EL-CodeBERT. The numerical value assigned to each item represents its respective p-value. The item value flagged in \textbf{bold} indicates that its p-value is less than 0.05. }}
	\label{tab:t-test}
	\resizebox{0.5\textwidth}{!}{%
		\begin{tabular}{@{}lllll@{}}
			\toprule
			{ Dataset} & ACC & P & R & F1 \\ \midrule

			{ Code Language } & 0.07676 & 0.57343 & 0.06107 & 0.10552\\
			{ Code Smell } & \textbf{0.00008} & \textbf{0.00007} & \textbf{0.00011} & \textbf{0.00011} \\
			{ Code Comment } & \textbf{0.00222} & 0.93387 & 0.08265 & 0.12453 \\
			{ Technical Debt } & 0.06752 & 0.55778 & 0.15295 & 0.06117 \\
			
			\bottomrule
		\end{tabular}%
	}
\end{table}

{To substantiate our findings, we have performed a  paired T-test comparing our results with those obtained from the EL-CodeBERT experiments.}  The reults are presented in Table \ref{tab:t-test}. It can be observed that, in the case of the dataset Code Language and Technical Debt, there is no discernible distinction between CodeClassPrompt and EL-CodeBERT. However, for the datasets Code Smell and Code Comment, notable differences exist. Referring to Table \ref{tab:codesmell}, it is evident that CodeClassPrompt outperforms EL-CodeBERT in all metrics pertaining to code smell classification. Moreover, as shown in Table \ref{tab:codecomment}, CodeClassPrompt demonstrates a slightly superior Accuracy metric in comparison to EL-CodeBERT for the code comment classification task.

\subsection*{Result Analysis for RQ2}

RQ2 aims to investigate the contribution of the attention and prompt components in our proposed approach. To demonstrate their contribution, we conducted extensive ablation experiments on four source code-related tasks, keeping the same experimental setup except for replacing the attention mechanism with a fully connected network or removing the prompt templates. Table \ref{tab:ablation} presents a comparison of the experimental results.
From the table, we observed that replacing the attention mechanism with a fully connected neural network results in worse performance compared to using the attention mechanism on all four tasks, particularly on programming language tasks, such as code classification and code smell classification. 
{
For the code comment classification and technical debt classification tasks, some individual evaluation metrics show slight increase, except for accuracy.}
After removing both the attention and prompt components, the model degenerates into a classifier based on fully connected feed-forward networks and CodeBERT, and its performance is worse on all metrics than CodeClassPrompt.

Comparing the last two rows in each task, we observed that utilizing the prompt template alone (denoted as "w/o Attention") does not improve performance in code smell classification (Accuracy value drops from 85.429\% to 82.286\%), and its contribution does not provide an explicit advantage in the other tasks. However, combining the prompt template with the attention mechanism can significantly enhance performance on all tasks. The results of our proposed CodeClassPrompt approach strongly demonstrate its effectiveness.

Comparing code language classification with code smell classification, we observed that the attention mechanism has a greater advantage over fully connected feed-forward networks (denoted as "w/o Attention\&Prompt") in code smell classification. The reason for this difference is that code smell requires the detection of semantic differences between code snippets of the same type, while code language classification involves detecting linguistic features of different programming languages. From the results, it appears that semantic features can be detected more easily than linguistic features.

\begin{table}[!ht]
	\centering
	\caption{Results of the ablation study. ``w/o attention" refers to the absence of the attention mechanism. The task marked with an asterisk (*) is a programming language task, while others are natural language tasks. ``w/o Attention\&Prompt" denotes the absence of both the attention mechanism and prompt templates, equivalent to a classifier based on fully connected feed-forward networks and CodeBERT. The values presented in \textbf{Bold} denote the optimal values achieved in each task. }
	\label{tab:ablation}
	{%
		\begin{tabular}{@{}llllll@{}}
			\toprule
			{Task} & { Method } &  { ACC }(\%) & {P}(\%) & {R}(\%) & {F1}(\%) \\
			\midrule
			\multirow{3}{*}{Code Language*}& \textsc{CodeClassPrompt} & \textbf{88.024} & \textbf{88.232} & \textbf{88.091} & \textbf{88.149} \\
			
			& {w/o Attention} & {87.828} & {88.010} & {87.926} & {87.958} \\
			& {w/o Attention\&Prompt}  & 87.418 & 88.042 & 87.450 & 87.614 \\
			\midrule
			\multirow{3}{*}{Code Smell*}& \textsc{CodeClassPrompt} & \textbf{86.000} & \textbf{86.167} & \textbf{85.657} & \textbf{85.824} \\
			& {w/o Attention} & {82.286} & {82.407} & {81.900} & {82.051} \\			
			&{w/o Attention\&Prompt } & 85.429 & 85.516 & 85.128 & 85.264 \\ 
			\midrule
			\multirow{3}{*}{Code Commet}& \textsc{CodeClassPrompt} & \textbf{95.416} & {89.654} & \textbf{87.036} & \textbf{87.930} \\
			& {w/o Attention} & {95.016} & \textbf{92.360} & {86.328} & {87.797} \\	
			&{w/o Attention\&Prompt } & 94.838 & 87.916 & 86.301 & 86.820 \\
			
			\midrule
			\multirow{3}{*}{Techical Debt}& \textsc{CodeClassPrompt} & \textbf{97.895} & \textbf{94.330} & {92.257} & \textbf{93.263} \\
			& {w/o Attention} & {97.745} & {92.677} & \textbf{93.337} & {93.004} \\
			&{w/o Attention\&Prompt } & 97.835 & 94.197 & 91.991 & 93.059 \\
			\bottomrule
			
		\end{tabular}%
	}
\end{table}

\noindent\textbf{Summary for RQ2:}
Our proposed CodeClassPrompt approach, which incorporates prompt templates and the attention mechanism to avoid additional computation costs associated with additional neural network layers, significantly improves performance on the four source code-related tasks compared to fully connected feed-forward networks. Both the attention and prompt components are valid and indispensable for achieving the best performance.





\subsection*{Result Analysis for RQ3}

\begin{figure*}
	\includegraphics[width=1.0\textwidth]{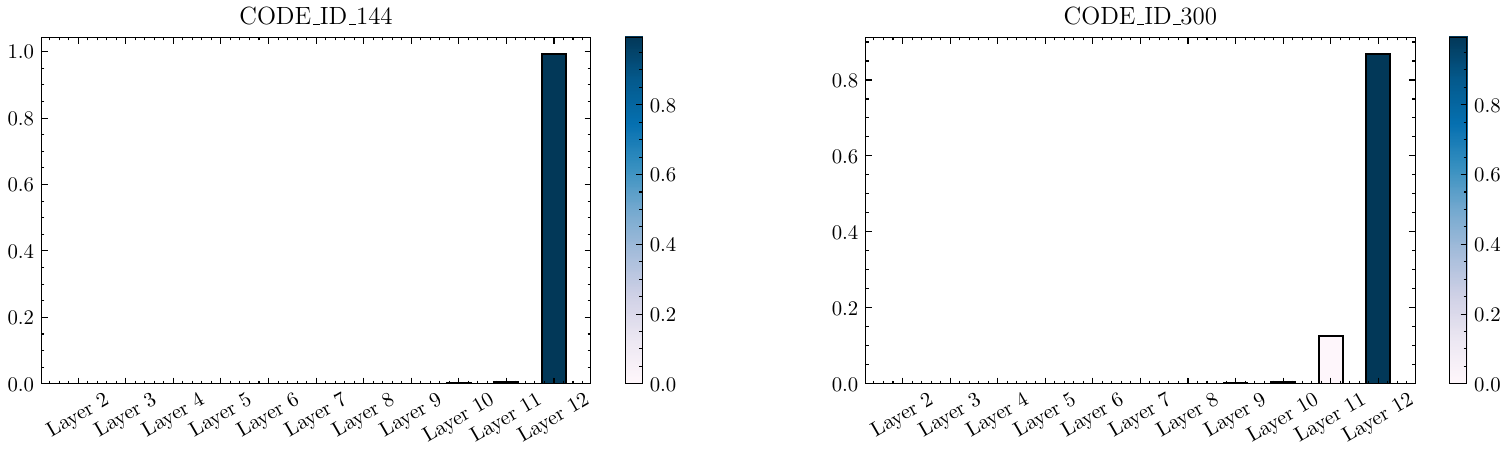}
	\caption{Attention values at each layer for code language classification}
	\label{fig:att-code}       
\end{figure*}

\begin{figure*}
	\includegraphics[width=1.0\textwidth]{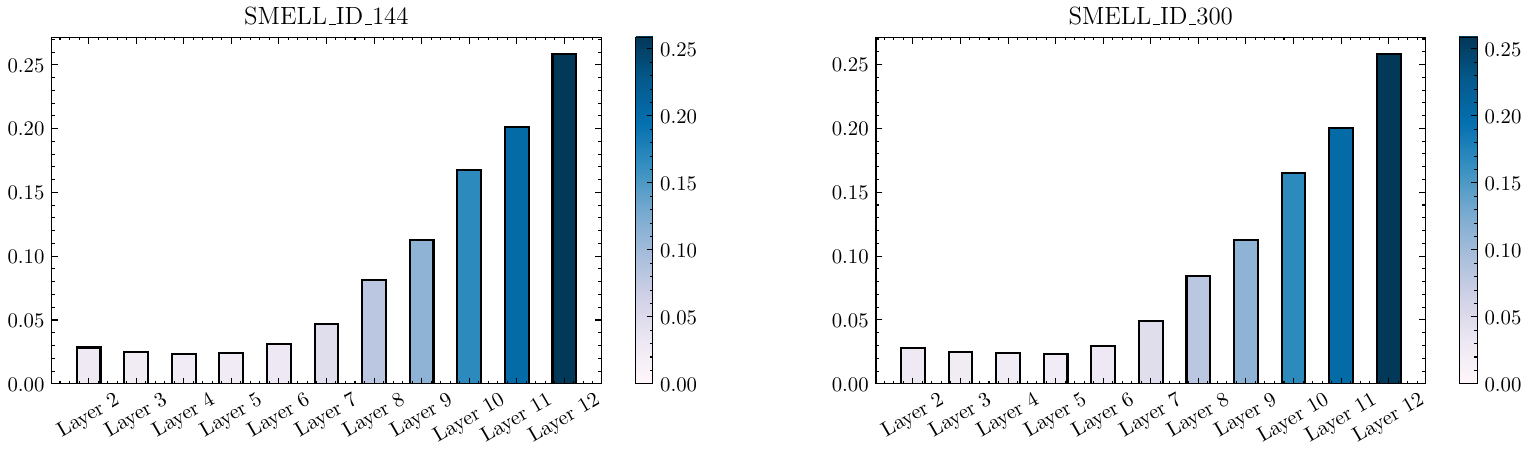}
	\caption{Attention values at each layer for code smell classification}
	\label{fig:att-smell}       
\end{figure*}

\begin{figure*}
	\includegraphics[width=1.0\textwidth]{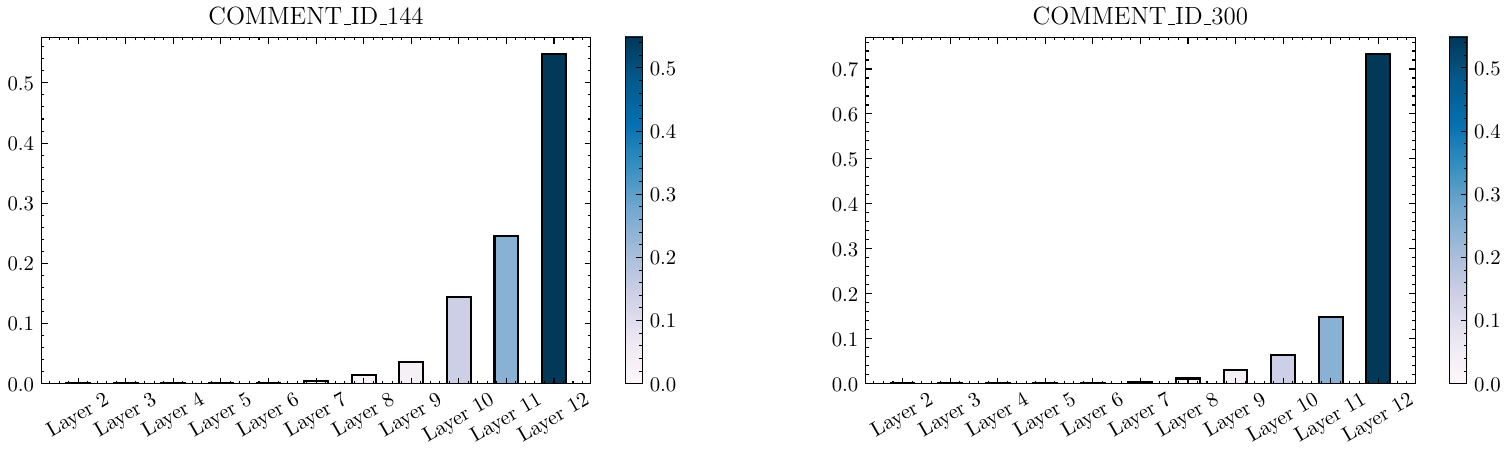}
	\caption{Attention values at each layer for code comment classification.}
	\label{fig:att-comment}       
\end{figure*}

\begin{figure*}
	\includegraphics[width=1.0\textwidth]{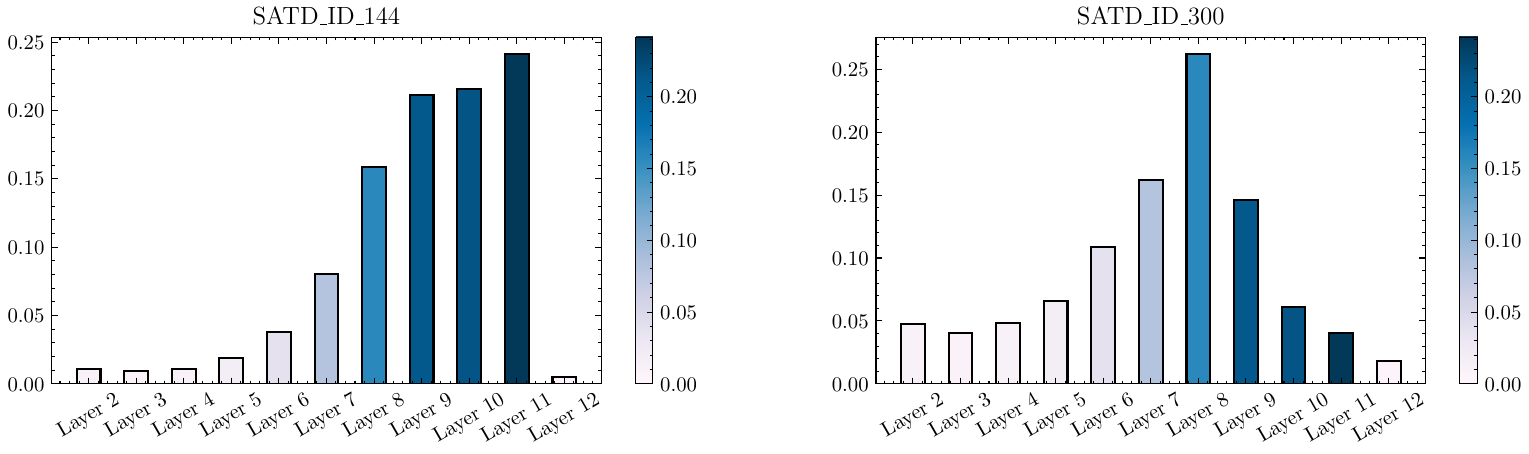}
	\caption{Attention values at each layer for technical debt classification}
	\label{fig:att-sadt}       
\end{figure*}

RQ3 aims to investigate the layers with decisive attention values that each task focuses on. We selected two examples from each task to examine the attention values of each layer. The details of each sample are lists in Table \ref{tab:attention}. The results are presented in Figures \ref{fig:att-code}, \ref{fig:att-smell}, \ref{fig:att-comment}, and \ref{fig:att-sadt}.

\noindent\textbf{Code Language Classification.}\quad As shown in Figure \ref{fig:att-code}, the concentration of attention in code language classification tasks is mainly on the last two layers of knowledge, particularly the final layer. For instance, in the case of id 144, the attention value focused on the last layer is 99.37\%, while that on layer 11 is only 0.42\%. Similarly, for id 300, attention on the last layers is 86.90\%, whereas that on layer 11 is 12.60\%. According to \cite{conneauWhatYouCan2018}, the last layer represents the highest level of knowledge regarding the input. As the code language classification task aims to distinguish the type of programming languages, the highest level of knowledge contains sufficient identifying features to accomplish this task.

\begin{table}[!ht]
	\centering
	\caption{ The text or code lists of attention values. ID denotes the id of a selected example.}
	\label{tab:attention}
	\resizebox{1.0\textwidth}{!}
	{%
		\begin{tabular}{@{}lll@{}}
			\toprule
			{Task} & {ID } &  { Text or Code Snippet }\\
			\midrule
			\multirow{2}{*}{Code Language}& \textsc{144} &  $\left \langle \%= Html.RadioButton("ticketStatus", "Open", true) \%  \right \rangle $  \\
			
			& \textsc{300} &  $Class DocumentVO{ PrintJobVO job; PrintRunVO run; String id; getters and setters.. }$\\
			\midrule								
			\multirow{2}{*}{Code Smell}&  \textsc{144} & $final int \left \langle w \right \rangle get Attributes  \left \langle /w \right \rangle  ( int id ) \{  \left \langle w\right \rangle ensure Id \left \langle /w\right \rangle \left( id \right) \dots \} $  \\
			
			& \textsc{300} & $ public void \left \langle w \right \rangle To Source \left \langle /w \right \rangle (  \left \langle w \right \rangle String Builder  \left \langle /w \right \rangle sb ) \{ String dot = \left \langle str\-literal \right \rangle \left \langle str-literal \right \rangle \dots \}$ ; \\
			\midrule
			\multirow{2}{*}{Code Comment}& \textsc{144} &  Contributors: * IBM Corporation - initial API and implementation * Bjorn Freeman-Benson\\
			
			& \textsc{300} &   // verify we are in main view and url is correct\\
			\midrule
			\multirow{2}{*}{Technial Debt}& \textsc{144} & //ChangeFactoryImpl \\
			
			& \textsc{300} &  ODO: What does the output directory have to do with the class path? Project p = ... ; \\
			\bottomrule
			
		\end{tabular}%
	}
\end{table}

\noindent\textbf{Code Smell Classification.}\quad As shown in Figure \ref{fig:att-smell}, code smell classification tasks focus on each layer of knowledge derived from a pre-trained model, but there is a higher concentration on the last few layers. For example, in the case of id 144, the attention values are 25.87\% on the final layer and 20.13\% on layer 11. The objective of code smell classification is to detect coding style and semantic information, and it relies on the features of knowledge that range from low to high levels.

\noindent\textbf{Code Comment Classification.}\quad As illustrated in Figure \ref{fig:att-comment}, this task primarily focuses on the last four layers of knowledge, with a stronger emphasis on the last three layers. For instance, for id 144, the attention values for layer 12, 11, and 10 are 54.84\%, 24.55\%, and 14.44\%, respectively. The last layer, which contains the highest-level knowledge, is particularly crucial for this task. Code comment classification is a natural language processing task that places a significant emphasis on differences in semantic information.

\noindent\textbf{Technical Debt Classification.}\quad As shown in Figure \ref{fig:att-sadt}, technical debt classification focuses more on the middle layers of knowledge rather than the last layers. For instance, in the case of id 144, it concentrates more on layer 7, 8, 9, 10, and 11, while for the last layer, the attention value is only 0.51\%. A similar trend can be observed for id 300. This suggests that the highest layer of knowledge contributes less to technical debt classification than other layers. This task is a binary classification task, similar to emotional classification, and does not involve concrete semantic information.

\noindent\textbf{Summary for RQ3:}  Various source code-related tasks demonstrate different degrees of attention on each layer of output knowledge from the CodeBERT model. This indicates that different levels of knowledge hold distinct meanings for different tasks, highlighting the effectiveness of the attention mechanism for source code-related tasks.



\begin{figure*}
	\centering\includegraphics[height=0.45\textheight]{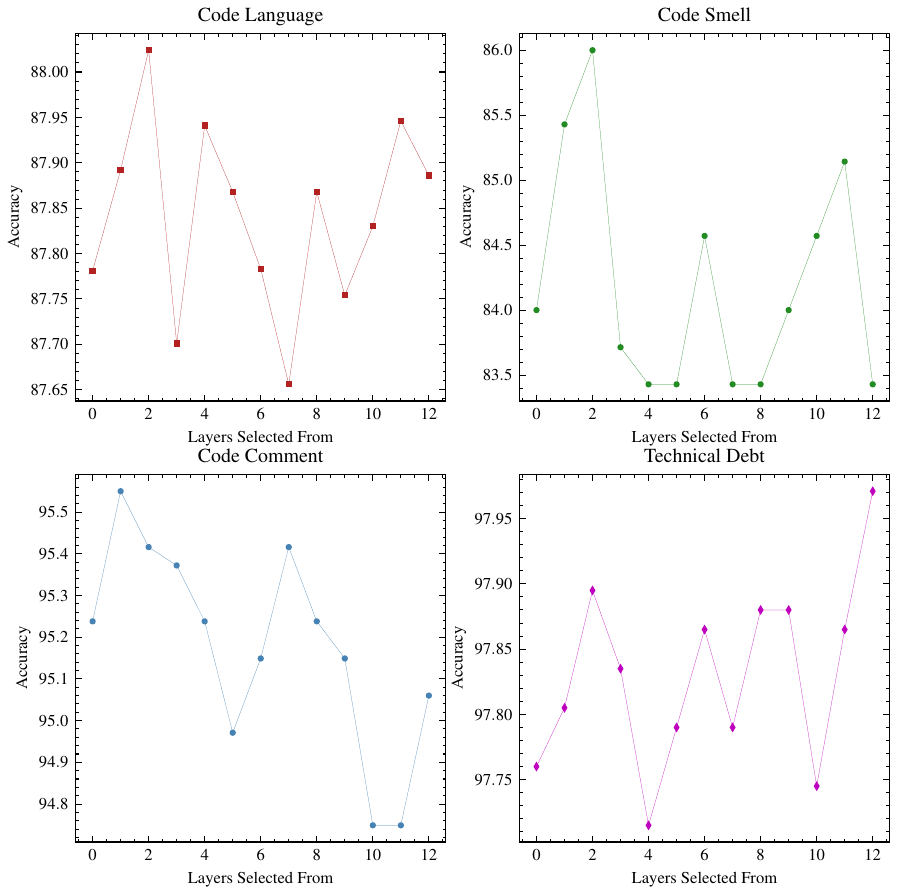}
	\caption{Results for the metric Accuracy} 
	\label{fig:accuracy}       
\end{figure*}

\begin{figure*}
	\centering \includegraphics[height=0.45\textheight]{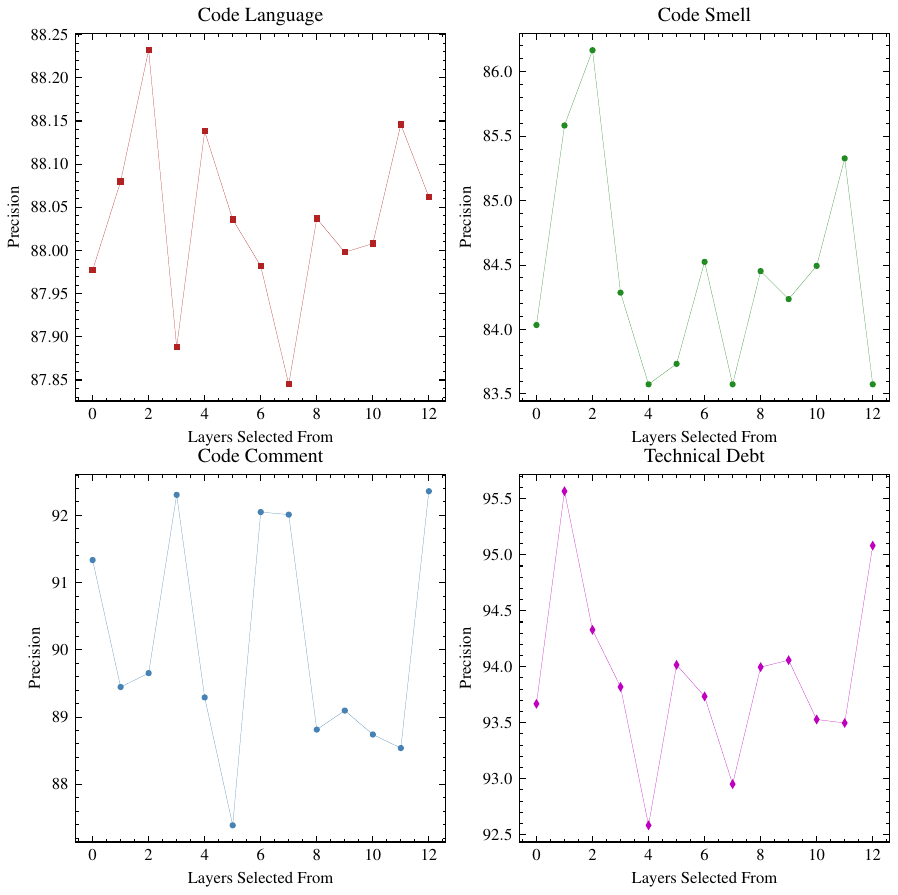}
	\caption{Results for the metric Precision}
	\label{fig:precision}       
\end{figure*}

\begin{figure*}
	\centering\includegraphics[height=0.45\textheight]{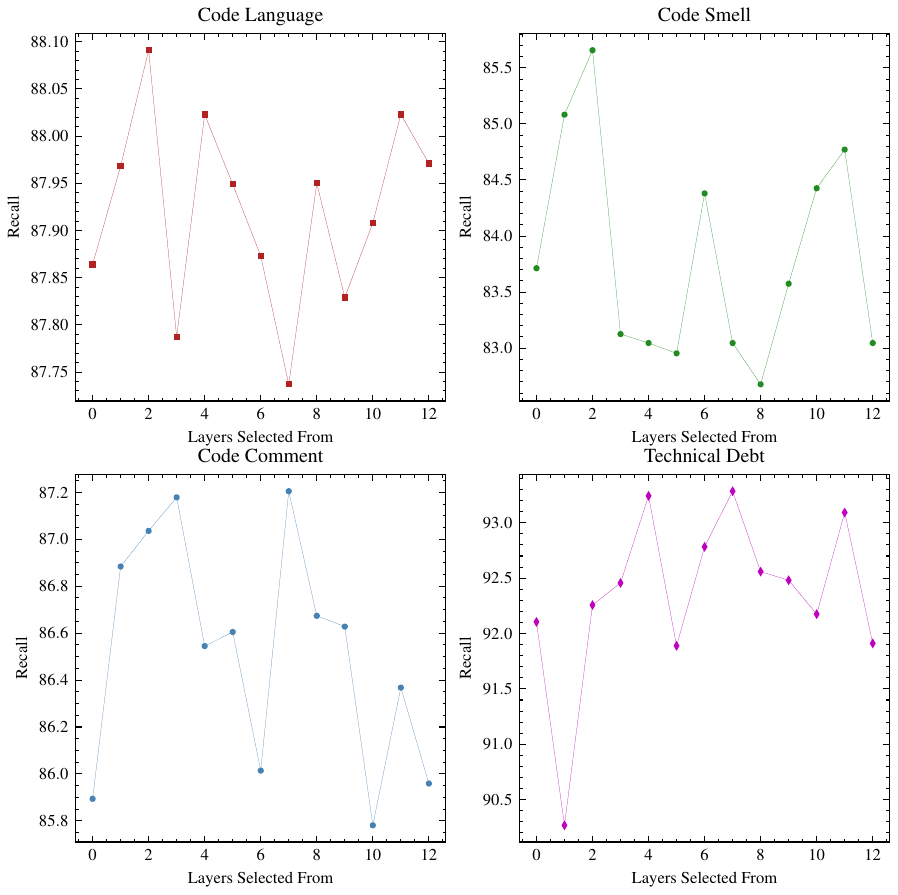}
	\caption{Results for the metric Recall}
	\label{fig:recall}       
\end{figure*}

\begin{figure*}
	\centering\includegraphics[height=0.45\textheight]{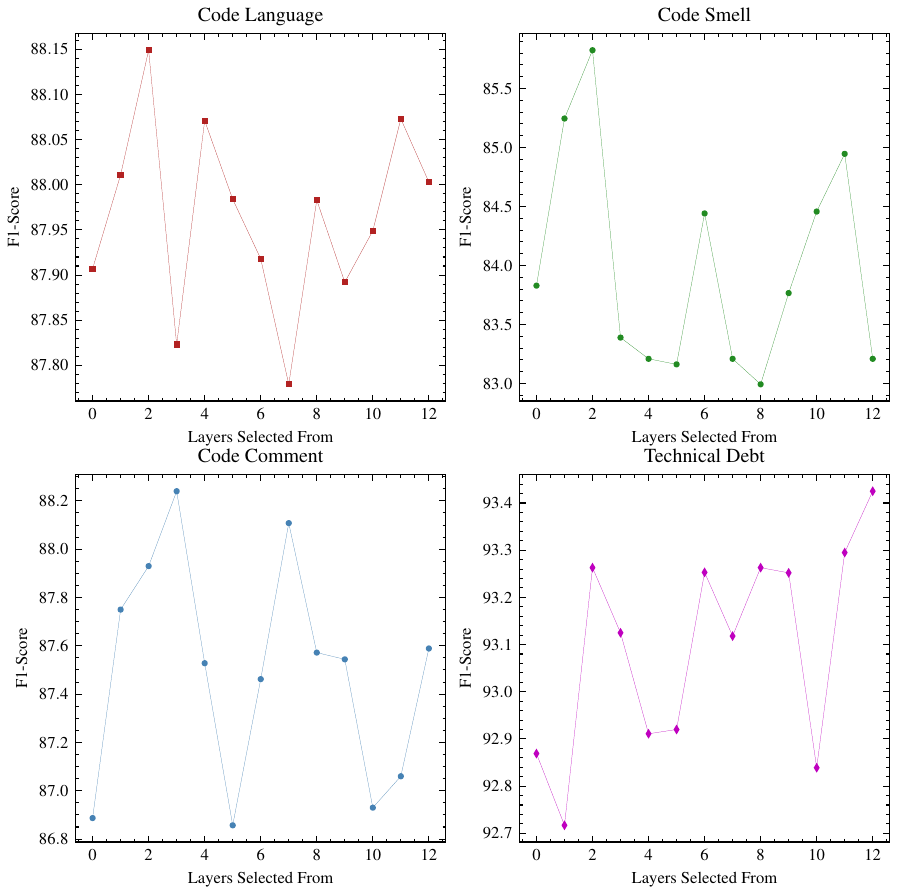}
	\caption{Results for the metric F1-score }
	\label{fig:f1-score}       
\end{figure*}

\subsection*{Result Analysis for RQ4}

The hidden state output of the BERT-based model consists of 13 layers (from layer 0 to layer 12), where layer 0 represents the embedding vector of the input information, and the other layers contain hierarchical linguistic features \cite{jawaharWhatDoesBERT2019}. {The primary objective of our experiments is to examine the influence of different layers of knowledge features on classification tasks related to source code. We aim to identify the most suitable layers that can serve as candidate layers for the attention mechanism, allowing us to aggregate crucial information effectively. } To this end, we collected 13 sets of knowledge features for each task, with the first set ranging from layer 0 through layer 12, the second set from layer 1 through layer 12, and so on. 

The four tasks can be grouped into two categories: programming language-based tasks (code language classification and code smell classification) and natural language-based tasks (code comment classification and technical debt classification). 
The experimental results for four metrics, namely Accuracy, Precision, Recall, and F1-score, are showcased in Figure \ref{fig:accuracy}, \ref{fig:precision}, \ref{fig:recall}, and  \ref{fig:f1-score} correspondingly.

{
From these figures, it is observed that the CodeClassPrompt approach consistently achieves the best results on the third set of knowledge features (from layer 2 through layer 12) across all evaluation metrics for programming language-based tasks. The attention mechanism is aimed to select the important features, the set with best results is regarded as the most important. For natural language processing tasks, our CodeClassPrompt approach on this set of features (from layer 2 through layer 12) still obtained better results than the baselines. These results demonstrate that the attention mechanism on knowledge features from layer 2 through layer 12 is better able to represent features on the programming language-based tasks. To ensure consistency across the four source code-related tasks, we have deliberately selected a fixed set of knowledge features, encompassing layers 2 through 12, as input candidates for the attention mechanism. By employing this specific range of layers, we aim to maintain uniformity in our approach across all tasks.
}

\noindent\textbf{Summary for RQ4:} Our CodeClassPrompt approach achieves greater performance on both programming language-related tasks and natural language-based tasks, with higher stability particularly for programming language-based source code-related tasks.

\subsection*{Result Analysis for RQ5}

\begin{table}[!ht]
	\centering
	\caption{Time consumption of an additional LSTM layer in CodeBERT based pipeline.}
	\label{tab:time_consumption}
	\resizebox{0.8\textwidth}{!}
	{%
		\begin{tabular}{@{}llllll@{}}
			\toprule
			{Task} & {Max Length } &{Repeat Times } &  { Total}(ms) & {LSTM}(ms) & {Percentage}(\%) \\
			\midrule
			\multirow{1}{*}{Code Language}& 256 & \textsc{10*1000} & \textsc{6430.27} & \textsc{719.15} & \textsc{11.18}  \\
			
			
			\bottomrule
			
		\end{tabular}%
	}
\end{table}

\begin{table}[!ht]
	\centering
	\caption{{Computation costs and parameters of CodeBERT based pipeline. "Comp Costs" refers to computational costs. "M" denotes one million (1,000,000). "GFLOPs" represents one billion (1,000,000,000) floating-point operations per second. The maximum sequence length is set to 256.  "CodeClassPrompt$_{\text{ }LSTM}$" refers to CodeClassPrompt with BiLSTM layers.}}
	\label{tab:computation_costs}
	\resizebox{0.8\textwidth}{!}
	{%
		\begin{tabular}{@{}lllll@{}}
			\toprule
			{Model} & {Parameters(M) } &{Comp Costs(GFLOPs) } &  { Reduced Parameters (\%)} & { Reduced Comp Costs (\%)}\\
			\midrule
			CodeClassPrompt$_{\text{ }LSTM}$& 109.87 & 22.05 & 0.0 & 0.0   \\
			{CodeClassPrompt}& 				\textsc{85.07}	& \textsc{21.76} & \textsc{22.57} & \textsc{1.32}   \\

			
			\bottomrule
			
		\end{tabular}%
	}
\end{table}

{
The computational efficiency experiment aimed to demonstrate the impact of additional neural network layers on a CodeBERT-based classification pipeline, including  parameter quantity, computatinal cost and time-saving\footnote{The computation costs and parameter numbers were evaluated using the Python library ``thop".}. Previous studies \cite{liuELCodeBertBetterExploiting2022} have utilized additional BiLSTM layers to extract more effective features and achieve notable performance with the CodeBERT-based pipeline. Using identical hardware and software configurations, including the GPU, CPU, memory, and software versions, as well as employing the same hyperparameters such as the maximum sequence length. 
{
	The computation cost of a specific BERT family model is solely dependent on the maximum length of the input content. For input content with a shorter length than the maximum, padding is applied to extend it to the specified maximum length. Conversely, if the input content exceeds the maximum length, truncation is utilized to ensure adherence to the predefined maximum length.
	} 
We selected an example from the dataset of code language classification and conducted ten groups of experiments, each consisting of 1,000 inference repetitions on the example. The resulting average time consumed by the ten groups is reported in Table \ref{tab:time_consumption}, where ``Total" denotes the time consumption of the entire pipeline, and ``LSTM" represents the time cost of the BiLSTM layer in the pipeline. The results show that the BiLSTM layer consumes 11.18\% of the time during the entire pipeline.
To assess the reasons behind the time-saving, we have analyzed the discrepancies in parameter numbers and computational costs between CodeClassPrompt and CodeClassPrompt with additional BiLSTM layers. The results are presented in Table \ref{tab:computation_costs}. From the table, it is evident that {the  number of parameters} is reduced by 22.57\%, while the computational costs are reduced by 1.32\%. By reducing the number of parameters, CodeClassPrompt effectively reduces memory access, leading to a significant decrease in the time required for transferring parameters from the CPU memory to the GPU memory. Additionally, the reduced computational effort further contributes to time savings. Collectively, these factors make CodeClassPrompt significantly more time-efficient.
}

\noindent\textbf{Summary for RQ5:} 
{
The experiments indicate that by leveraging powerful feature extraction capabilities, the removal of an additional neural network layer eliminates redundant parameters and computational costs, thereby substantially reducing computation time.
}

\section*{{Discussion}}
\subsection*{{Error Analysis}}
The following analysis pertains to the meticulous examination of classification errors within datasets that primarily focus on source code classification. These datasets encompass two distinct tasks: code language classification and code smell classification.
\subsubsection*{{Code Languae Classification}}

{By conducting a comprehensive investigation of error examples, we find that they can be classified into seven distinct groups.} The detailed categories and corresponding samples are provided in Table \ref{tab:error_code_lang}. The below part presents an in-depth analysis of the seven identified groups:

\begin{itemize}
	\item \textbf{Short} \quad 
	
	The content provided is insufficient in length to accurately determine its class information.
	\item \textbf{Pseudo} \quad 
	
	The information provided appears to resemble pseudo-code, making it difficult to detect its specific class or category.
	\item \textbf{Non-code}\quad
	
	 The provided content clearly does not resemble any form of source code or pseudo-code.
	\item \textbf{No-feature}\quad 
	
	{The examined source code displays characteristics that align with multiple code languages. Specifically, the analyzed code snippet lacks the distinctive language-specific features associated with a single programming language.}
	\item \textbf{ErrorClass}\quad 
	
	Despite the presence of evident features in the code snippet, the classifier mistakenly assigns it to an incorrect class.
	\item \textbf{LikeButNone}\quad 
	
	Although the content bears resemblance to a certain type of source code, it is, in fact, not a valid representation of source code.
	\item \textbf{Mix}\quad 
	
	The given code snippet exhibits characteristics of multiple programming languages, leading to a mixed representation. As a consequence, the classifier may encounter difficulties in accurately categorizing the content.
\end{itemize}

\begin{table}[!ht]
	\centering
	\caption{Examples of error classification in code language.}
	\label{tab:error_code_lang}
	\resizebox{0.99\textwidth}{!}
	{%
		\begin{tabular}{@{}ll@{}}
			\toprule
			{Category} & {Examples}  \\
			\midrule
			\multirow{2}{*}{Short}&  BitSet ++= \\
									& retUnique() \\
			\midrule
			\multirow{2}{*}{Pseudo}&  <if>abc <else>xyz <if> <else> <if> <else> abc xyz \\
				& |Type Object pointer| | Sync Block | | Instance fields...| | Instance fields...| \\

			\midrule
			\multirow{2}{*}{Non-code}&  www.domain.com/first/second/last/ last www.domain.com/last/ www.domain.com/first/second/third/fourth/last ... \\
							& abc1 abc2 abc3 abc1 ok abc2 ok abc3 ok \\
			\midrule
			
			\multirow{2}{*}{No-feature}& f a b = ((a+b) == 2) \&\& ((a*b) == 2) \&\& \\
										& \$x = \$hash{blah} || 'default' \\
			\midrule
			\multirow{2}{*}{ErrorClass}& std::string mystr="MY-PC" bSuccess = SetComputerNameA(mystr.c\_str()); if( bSuccess == 0 ) printf\"Unable to change computer name ...  \\
									& <?php \$this->dojo()->setLocalPath(\$this->baseUrl().'/javascript/dojo/dojo.js') ...\\
				\midrule
			\multirow{2}{*}{LikeButNone}&  <Return Address> \\
									& <function appendNextFib at 0x01FB14B0> \\
				\midrule			
			\multirow{2}{*}{Mix}&  http://server/base/feeds/documents?bq=[type in {'news'}] bq=[type = 'news'] -> return ["news"] bq=[type in {'news'}] -> return ["news"] ...			 \\
							& < script type="text/javascript" src="js/jquery.query-2.1.6.js"> </script> <? \$next\_exp = 123; ?> \$(document).ready(function() ... \\

			\bottomrule
			
		\end{tabular}%
	}
\end{table}

After conducting a comprehensive analysis of all the incorrect cases, it has been observed that the "None-code" type of snippets constitutes almost half of the total, closely followed by the "Short" type.

\subsubsection*{{Code Smell Classification}}
Code smell classification is a binary classification task, where the errors can be categorized into three distinct types. The detailed breakdown of these types is provided in Table \ref{tab:error_code_smell}. Next, we will provide detailed descriptions of these types.

\begin{itemize}
	\item \textbf{Short} \quad 
	
		The provided code snippet is of insufficient length to reliably determine its class information accurately.
	\item \textbf{Bad-label} \quad 
	
		 Despite the dataset assigning an incorrect label to the given code snippet, the classifier successfully identified its correct type.
	\item \textbf{Error-class}\quad
	
		The trained classifier incorrectly assigns an erroneous label to the code snippet.

\end{itemize}

After conducting a comprehensive analysis of all the erroneous cases, it has been observed that the "Short" type of snippets constitutes more than one-quarter of the total. Assessing the quality or undesirability of a snippet that is excessively brief presents considerable difficulty.

Based on the aforementioned discussion, it is evident that the primary drawback stems from the suboptimal quality of the datasets, primarily caused by the introduction of noisy data.

\begin{table}[!ht]
	\centering
	\caption{Examples of error classification in code smell.}
	\label{tab:error_code_smell}
	\resizebox{0.6\textwidth}{!}
	{%
		\begin{tabular}{@{}ll@{}}
			\toprule
			{Category} & {Examples}  \\
			\midrule
			\multirow{2}{*}{Short}&  final int [ ] stack ; \\
			& private String xtends ; \\
			\midrule
			\multirow{1}{*}{Bad-label}& <comment> private String <w> non Proxy Hosts </w>  \\

			\midrule
			\multirow{1}{*}{Error-class}&  <comment> private int <w> skipped Positions </w> ; \\

			\bottomrule
			
		\end{tabular}%
	}
\end{table}


\subsection*{{Further Investigation}}

This work aims to enhance computational efficiency by leveraging the prompt-learning paradigm. The effectiveness of manually defined templates has been demonstrated in source code-related classification tasks. However, it is worth examining whether automatically constructed templates can outperform manual prompt templates in source code-related tasks. Furthermore, while zero-shot based prompt learning has proven effective in natural language processing tasks, its applicability to source code-related tasks, particularly those involving programming languages, requires further investigation.

\section*{Threats to Validity}

In this section, we will focus on discussing potential threats to the validity of our empirical study.

\subsection*{Interanl Threats}
{
The internal validity threats to our research are predominantly associated with the experimental milieu. The initial concern pertains to the selection of hardware and software platforms, which may influence the reliability of our method's execution. To ameliorate this issue, we ensured a consistent experimental framework by employing uniform hardware configurations and opting for established software versions, including PyTorch and the Linux operating system. The second threat is rooted in the execution of baselines, which we addressed by sourcing code from reputable, well-established libraries. Finally, the third threat involves the stochastic nature of deep learning model initializations. To guarantee the reproducibility of our findings, we utilized fixed random seeds across all experimental trials.
}

\subsection*{External Threats}

The main external threat lies in the choice of datasets used for the four downstream tasks related to source code. To mitigate this threat, we have opted to use publicly available corpora. Specifically, for code language classification, we draw upon the dataset provided by Alrashedy et al. \cite{alrashedySCCPredictingProgramming2020a}. For code smell classification, we utilize the dataset from Fakhoury et al. \cite{fakhouryKeepItSimple2018}. For code comment classification, we rely on the dataset prepared by Pascarella et al. \cite{pascarellaClassifyingCodeComments2017}. Lastly, for technical debt classification, we employ the dataset introduced by Maldonado et al. \cite{maldonadoUsingNaturalLanguage2017,fakhouryKeepItSimple2018}.

\subsection*{Construct Threats}

The primary constructive threat we address in this study is the selection of appropriate evaluation metrics for assessing performance on source code-related tasks. To ensure a fair and comprehensive comparison, we have selected four widely-used metrics (accuracy, precision, recall, and F1-Score) that have been extensively employed in previous studies, such as the work by Alrashedy et al. \cite{alrashedySCCPredictingProgramming2020a}.

\section*{Conclusion}
{
{
In this study, we introduced CodeClassPrompt, a novel approach that harnesses relevant knowledge extracted from a pre-trained model to improve source code-related classification. Through empirical analysis, we demonstrated the effectiveness of CodeClassPrompt, achieving enhanced computational efficiency while yielding comparable results to previous studies. CodeClassPrompt consolidates multi-layer knowledge into a unified input representation, eliminating the need for additional neural layers and thereby reducing computational costs. This efficiency is further supported by computational cost experiments, while ablation studies validate the utility of the multi-layer attention mechanism in combination with the prompt learning paradigm. Additionally, attention analysis provides insights into the distinct contributions of each layer of the output from pre-trained language models (PLMs) to the performance of various tasks.
}


\bibliography{promptlearning.bib,codeprompt.bib}

\section*{Acknowledgements}

The work was supported by the 242nd National Information Security Project (No. 2020A065).

\section*{Author contributions statement}

Y.M. developed the study concept. Y.M. and Y.S conceived the experiment(s),  Y.M. and Y.S. conducted the experiment(s), Z.L., S.L. and Y.S. analysed the results. The final manuscript was written by Y.M., Y.Z. All authors reviewed the manuscript.

\section*{Conflict of interest}
The authors declared that they have no conflict of interest.

\section*{Data Availability Statements}

The data supporting the findings of this study are openly available in the GitHub repository at https://github.com/BITENGD/codeclassprompt.

\end{document}